\pdfoutput=1
\documentclass[10pt, logo, intern]{template/nvidiatechreport}

\usepackage{natbib}
\usepackage{hyperref}

\usepackage{url}
\usepackage{xspace}
\usepackage{amsmath}
\usepackage{amssymb}
\usepackage{graphicx}
\usepackage[dvipsnames]{xcolor}
\usepackage{listings}
\usepackage{multirow}
\usepackage{rotating}
\usepackage{subcaption}
\usepackage{tabularx}
\usepackage{booktabs}
\usepackage{colortbl}
\usepackage{comment}
\usepackage{fontawesome5}
\usepackage[noabbrev,capitalise]{cleveref}

\crefname{lstlisting}{listing}{listings}

\definecolor{githubgray}{RGB}{240,240,240}      
\definecolor{githubblue}{RGB}{0,92,197}
\definecolor{githubgreen}{RGB}{22,163,74}
\definecolor{githubred}{RGB}{207,34,46}
\definecolor{githubpurple}{RGB}{136,46,224}
\lstdefinestyle{github}{
    language=Python,
    basicstyle=\ttfamily\small,             
    keywordstyle=\color{githubpurple},           
    commentstyle=\color{githubgreen}\itshape,    
    stringstyle=\color{githubred},               
    numberstyle=\tiny\color{gray!60},            
    numbers=left,
    stepnumber=1,
    numbersep=8pt,                              
    showstringspaces=false,
    breaklines=true,
    frame=none,                                  
    backgroundcolor=\color{githubgray},          
    xleftmargin=0pt,                            
    framexleftmargin=0pt,
    rulecolor=\color{gray!30},
    tabsize=2,                                  
    captionpos=b,
    basewidth=0.5em,                            
    columns=fixed
}

\newcommand{\kvcache}{KV cache\xspace}
\newcommand{\ours}{Expected Attention\xspace}

\newcommand{\deeps}{Qwen-15B-R1\xspace}
\newcommand{\deepsb}{Qwen-7B-R1\xspace}
\newcommand{\llama}{Llama3.1-8B\xspace}
\newcommand{\gemma}{Gemma3-12B\xspace}
\newcommand{\qwen}{Qwen3-8B\xspace}

\newcommand{\omath}{OpenMath-Nemotron-14B\xspace}

\newcommand{\m}[1]{\mathbf{#1}}

\definecolor{darkblue}{rgb}{0, 0, 0.3}
\hypersetup{colorlinks=true, citecolor=darkblue, linkcolor=darkblue, urlcolor=darkblue}

\definecolor{githubblue}{HTML}{0000FF}
\definecolor{hforange}{HTML}{FF6600}
\definecolor{nvgreen}{HTML}{76B900}

\newcommand{\githubrepo}[1]{\href{#1}{{\faGithub} NVIDIA/KVPress}}

\newcommand{\hflink}[2]{%
  \href{#1}{\raisebox{-0.9ex}{\includegraphics[height=1.5em]{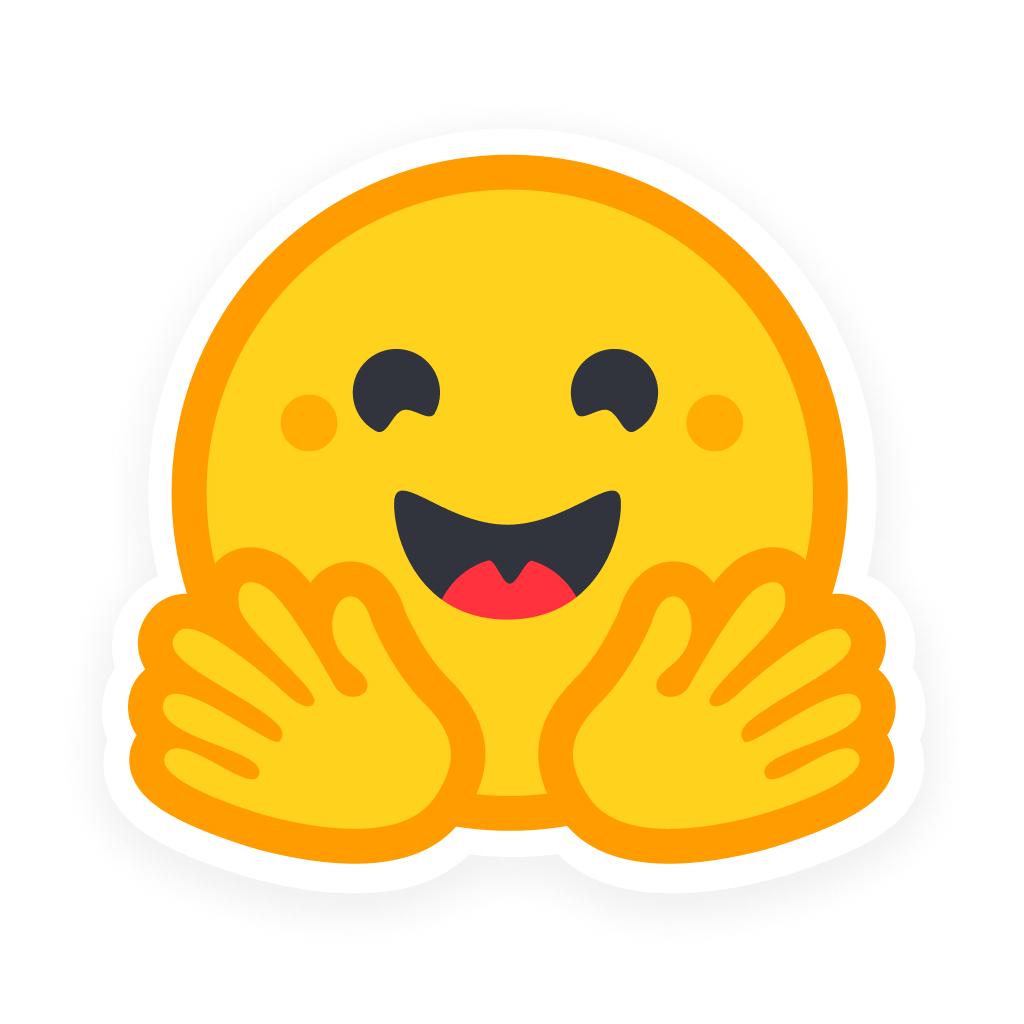}} #2}%
}

\title{Expected Attention:\\ KV Cache Compression by Estimating Attention from Future Queries Distribution}

\newcommand{\nvidia}{$^\dagger$}
\newcommand{\sapienza}{$^\diamond$}
\newcommand{\intern}{$^*$}

\author{ Alessio Devoto\intern\sapienza~~~~~Maximilian Jeblick\nvidia~~~~~~Simon Jégou\nvidia \\
  \sapienza Sapienza University of Rome~~~~\nvidia NVIDIA\\
  \begin{center}
    \githubrepo{https://github.com/NVIDIA/kvpress} \quad
    \hflink{https://huggingface.co/spaces/nvidia/kvpress-leaderboard}{KVPress Leaderboard}
\end{center}
}
\begin{abstract}
Memory consumption of the Key-Value (KV) cache represents a major bottleneck for efficient large language model (LLM) inference. While attention-score-based KV cache pruning shows promise, it faces critical practical limitations: attention scores from future tokens are unavailable during compression, and modern implementations like Flash Attention do not materialize the full attention matrix, making past scores inaccessible. To overcome these challenges, we introduce \textbf{Expected Attention, a training-free compression method} that estimates KV pairs importance by predicting how future queries will attend to them. Our approach leverages the distributional properties of LLM activations to compute expected attention scores in closed form for each KV pair. These scores enable principled ranking and pruning of KV pairs with minimal impact on the residual stream, achieving effective compression without performance degradation. Importantly, our method operates seamlessly across both prefilling and decoding phases, consistently outperforming state-of-the-art baselines in both scenarios. Finally, \textbf{we release KVPress, a comprehensive library to enable researchers to implement and benchmark \kvcache compression methods, already including more than 20 techniques.}
\end{abstract}

\begin{document}

\maketitle

\section{Introduction}
Large language models (LLMs)~\citep{gpt4, claude, llama4, qwen3} have revolutionized text generation and reasoning, enabling advanced applications such as long multi-round dialogues, extensive multimodal intelligence~\citep{qwen3, weng2024longvlm}, and agentic workflows that ingest massive amounts of data~\citep{openai_o1_2024, perpdeepresearch, yamada2025aiscientistv2workshoplevelautomated}. These applications often require processing extensive contextual information. For example, processing a large codebase or a short video can easily involve analyzing hundreds of thousands of tokens. A critical issue in deploying LLMs in such scenarios is the prohibitive memory consumption of the Key-Value (KV) cache~\citep{yaofu-long-context-challenge, kvreview, kvsurvey}.

During autoregressive generation, the \kvcache stores key and value vectors for every processed token, enabling efficient attention computation. However, its memory footprint grows linearly with sequence length, quickly becoming the primary bottleneck for long-context inference. A medium-sized 70B model~\citep{llama3} requires approximately 320 GB of GPU memory for a one-million-token \kvcache, far exceeding most GPU capacities. This challenge intensifies with emerging applications where advanced reasoning models generate thousands of intermediate tokens~\citep{deepseekv3, qwen3} and agentic systems load massive datasets~\citep{oai2025deepresearch,perpdeepresearch}. While current LLMs promise extended context lengths up to a million tokens~\citep{comanici2025gemini25pushingfrontier, llama4}, hardware constraints saturate GPU memory well before reaching theoretical limits.

State Space Models offer a solution by reducing memory costs~\citep{ssm, mamba}, yet their inferior performance compared to transformers, especially on long context tasks, limits adoption~\citep{repeat, illusionof}. Other architectural changes limited to the attention mechanism, such as multi-head latent attention~\citep{deepseekv2} or sliding window attention~\citep{mistral, gemma_2025}, reduce \kvcache size but do not remove the attention bottleneck and are orthogonal to \kvcache compression. Additionally, such methods need to be implemented at training time, limiting their application to pre-trained modern LLMs. This creates demand for training-free \kvcache compression methods that preserve transformer architectures while mitigating memory growth.

A promising direction for such training-free compression lies in exploiting semantic redundancy in natural language: not all tokens equally influence future predictions, and many provide negligible information once their contextual role is fulfilled. This property allows to compress the \kvcache by removing some of the keys and values stored in it. However, determining which tokens can be safely removed is far from trivial, as any Key-Value (KV) pair's importance depends on how \textit{future queries} will attend to it. Existing approaches use heuristics like discarding oldest tokens~\citep{fastgen, streamingllm} or leverage attention scores from past queries~\citep{h2o, snapkv, tova}, but these strategies are limited for real-world scenarios, and often require accessing attention scores which are not materialized in modern transformer implementations~\citep{flashattention}.

Instead of relying on heuristics or local attention metrics, we argue that a KV pair's significance is best measured by its global effect on the transformer's output. We quantify this effect by isolating each KV pair's contribution within the residual stream, capturing its influence on the model output. This raises the challenge of estimating \textit{how future queries will attend to each token in the context}, which requires accessing attention scores from the past and from future tokens, that are not available at the time of compression. To address this, we introduce \textit{\ours}, which estimates future attention allocation from distribution of future queries. \ours estimates the importance that each token in the context has for queries that have not been generated and accordingly prunes the \kvcache up to 60\% while preserving performance quality, requiring no architectural modifications or additional training. We release our code as a comprehensive library benchmarking over 20 state-of-the-art compression methods.

To summarize, our contributions are the following:
\begin{itemize}
    \item We analyse the distributional properties of LLM activations through the lenses of \kvcache compression and introduce the concept of \textit{\ours} to estimate the importance that current tokens will have in the future.
    \item We introduce a \kvcache compression method that leverages \ours and evicts irrelevant KV pairs for efficient inference. 
    \item We release all our code as a library, designed for researchers, that allows to easily implement, test and benchmark \kvcache compression methods.
\end{itemize}
\begin{figure}[b]
    \centering
    \begin{subfigure}[b]{0.49\linewidth}
        \centering
        \includegraphics[width=\linewidth]{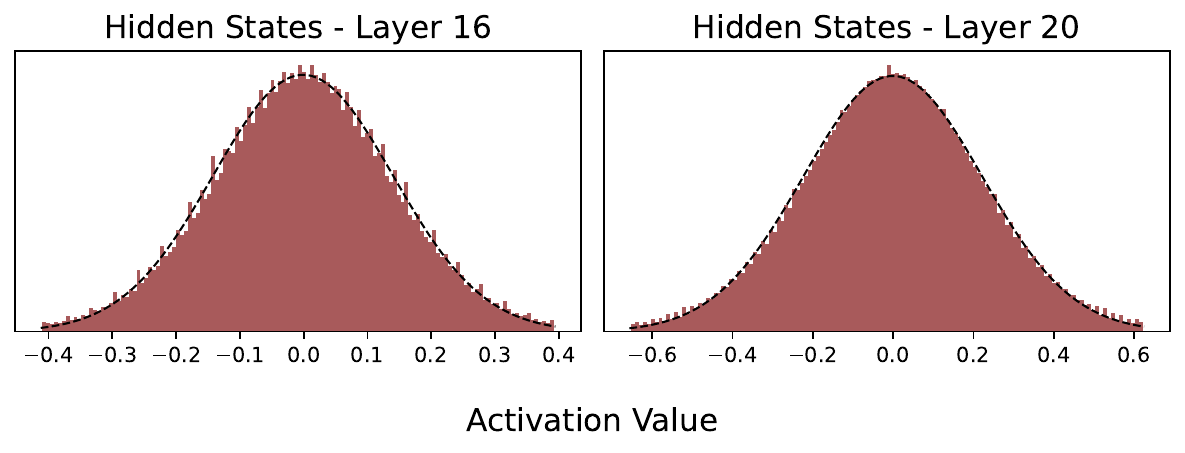}
    \end{subfigure}
    \hfill
    \begin{subfigure}[b]{0.49\linewidth}
        \centering
        \includegraphics[width=\linewidth]{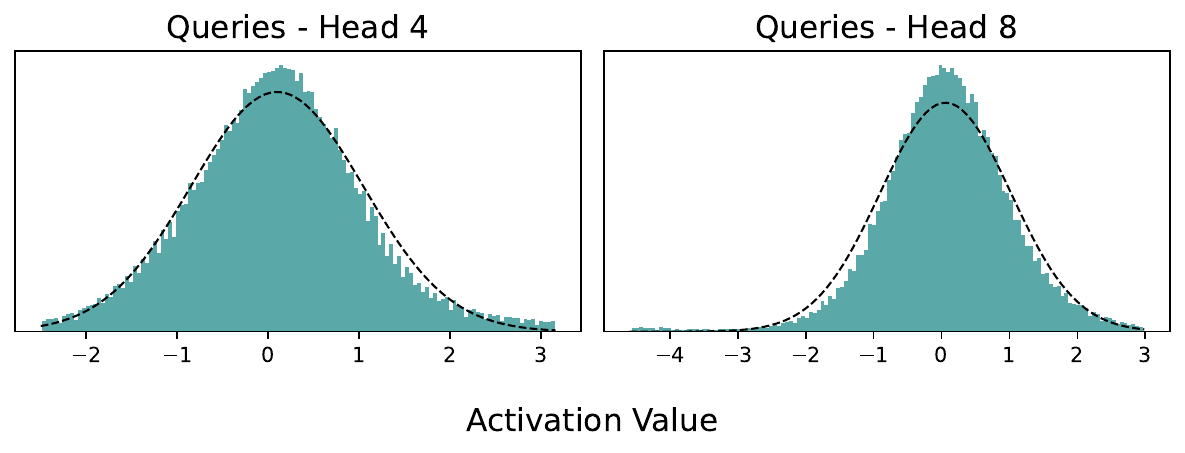}
    \end{subfigure}
    \caption{Hidden states from layer 16 and 20 and corresponding queries for layer 20 in \llama. Hidden states in modern LLMs are mostly normally distributed. As a consequence, query activations also follow a Normal. The best Gaussian fit is overlayed. We show more examples and discuss this property in \Cref{app:adistributions}.}
    \label{fig:gaussian_hs}
\end{figure}
\section{Expected Attention}
\label{sec:expected_attention}
\subsection{Key-Value Cache in Autoregressive Transformers} 
We consider decoder-only language models based on the transformer architecture~\citep{transformer}, representing the vast majority of modern LLMs. When an input sequence of tokens $\mathbf{x} = [x_1, x_2, \ldots, x_t]$ is fed to the model, each token $x_i$ is transformed into a hidden state representation $h_i \in \mathbb{R}^{h}$ and processed by a stack of transformer layers, including feed forward networks and multi-head attention blocks. For brevity and clarity, we focus our analysis on a single layer and attention head, noting that the following analysis naturally extends to multi-head attention, grouped query attention (GQA, \citealt{mqa}) and all their variants.

Let $h_i \in \mathbb{R}^{h}$ denote the hidden state at position $i$ in the sequence. In the attention block, the corresponding Query, Key and Value projections are computed as:
\begin{equation}
\label{eq:proj}
q_i = R_i W_Q h_i, \quad k_i = R_i W_K h_i, \quad v_i = W_V h_i
\end{equation}
where $d$ is the attention head dimension, $R_i \in \mathbb{R}^{d \times d}$ is the Rotary Position Embedding (RoPE, \citealt{rope}) matrix at position $i$, and $W_Q, W_K, W_V \in \mathbb{R}^{h \times d}$ are respectively the learnable projection matrices for query, key, and value in $\mathbb{R}^{d}$. During autoregressive inference, keys and values vectors are stored in the \kvcache to avoid recomputing them in future generation steps. The resulting \kvcache is a collection of Key-Value pairs $(k_i, v_i)$ from all inference steps in the sequence, leading to significant computational savings but increasing memory requirements, growing linearly with sequence length.

At generation step $t$, the attention mechanism computes the attention score between the current query $q_t$ and each previously cached key $k_i$ for $i \leq t$:
\begin{equation}
\label{eq:attn_score}
    a_{ti} = \frac{\exp\left( \frac{q_t^T k_i}{\sqrt{d}} \right)}{\sum_{j=1}^t \exp\left( \frac{q_t^T k_j}{\sqrt{d}} \right)} = \frac{z_{ti}}{\sum_{j=1}^t z_{tj}}
\end{equation}
where $a_{ti}$ is the normalized attention score between query at position $t$ and key at position $i$, and $z_{ti} = \exp\left( \frac{q_t^T k_i}{\sqrt{d}} \right)$ represents the unnormalized attention score. 

The attention score is used to weight and sum over all values previously stored in the \kvcache. The resulting output is then added to the hidden state $h_t$:
\begin{equation}
\label{eq:attention_update}
    h_t^{\text{out}} = h_t + \sum_{i=1}^t a_{ti} W_o v_i =  h_t + \sum_{i=1}^t \Delta h_{ti}
\end{equation}
where $h_t \in \mathbb{R}^h$ and $h_t^{\text{out}} \in \mathbb{R}^h$ represent the hidden state before and after the attention update respectively, and $W_o \in \mathbb{R}^{d \times h}$ is the learnable output projection matrix.  The hidden states embedding $h_t$ represents the "residual stream,"~\citep{elhage2021mathematical} updated via vector additions by each transformer block. The value $\Delta h_{ti} = a_{ti} W_o v_i$ isolates the specific residual addition of the $i$-th KV pair at step $t$. This decomposition reveals that each cached KV pair $(k_i, v_i)$ contributes a residual update $\Delta h_{ti}$ to the final output, and provides a natural measure of the importance of each KV pair:
\begin{equation}
\label{eq:score}
    \|\Delta h_{ti}\| = a_{ti} \|W_o v_i\|
\end{equation}
where $\|\cdot\|$ denotes the L2 norm. This metric captures both the attention weight $a_{ti}$ (how much the query attends to the $i$-th key) and the transformed value magnitude $\|W_o v_i\|$ (the impact of the $i$-th value on the output).
Equation~\ref{eq:score} provides the optimal measure for estimating the impact of each KV pair in the model output. If we could compute this score for all cached KV pairs, we could selectively prune the cache by removing pairs with the lowest impact, thereby minimizing performance degradation. However, computing Equation~\ref{eq:score} presents significant practical challenges. While $\|W_o v_i\|$ is readily available at inference time, the attention weight $a_{ti}$ depends on future queries that have not yet been generated. Specifically, we cannot know the attention scores from future tokens $t+1, t+2, \ldots$ before computing them, making it impossible to predict which KV pairs will be important for upcoming generation steps. Furthermore, modern transformer implementations utilize Flash Attention~\citep{flashattention, flashattention2}, which computes attention scores on-the-fly without materializing the complete attention matrix, preventing access to even past attention scores. To address these fundamental limitations, we leverage the properties of activations in modern LLMs, and introduce \textit{\ours}.
\subsection{Expected Attention: Estimating Attention From Future Queries}
\paragraph{Distributional properties of LLM activations}
To approximate the unnormalized attention score $z_{ij}$,  we leverage the findings of ~\citet{teal}, showing that hidden states in modern LLMs loosely follow a Gaussian distribution $h \sim \mathcal{N}(\mu, \Sigma)$. While we show an example of this property in \Cref{fig:gaussian_hs}, we also extensively validate it across multiple model architectures in \Cref{app:adistributions}. Given this distributional assumption, queries also inherit Gaussian properties through the linear transformation in Equation~\ref{eq:proj} $q_t = R_t W_Q h_t$:
\begin{equation}
    q_t \sim \mathcal{N}(\mu_{q_t}, \Sigma_{q_t}), \quad \text{where } \mu_{q_t} = R_t W_Q \mu, \quad \Sigma_{q_t} = R_t W_Q \Sigma W_Q^T R_t^T
\end{equation}
where $\mu \in \mathbb{R}^d$ and $\Sigma \in \mathbb{R}^{d \times d}$ are the mean and covariance of the hidden state distribution, and $R_t \in \mathbb{R}^{d \times d}$ is the RoPE matrix at position $t$.

To create a single, tractable representation of attention over a future interval, we approximate the positional embeddings by averaging the RoPE matrix over the next $T$ positions.  This gives us a position-averaged query distribution:
$
$
\begin{equation}
    \bar{q} \sim \mathcal{N}(\bar{\mu}_q, \bar{\Sigma}_q), \quad \text{where } \bar{\mu}_q = \bar{R} W_Q \mu, \quad \bar{\Sigma}_q = \bar{R} W_Q \Sigma W_Q^T \bar{R}^T
\end{equation}
$
$
where $\bar{R} = \frac{1}{T} \sum_{j=1}^T R_{t+j}$ represents the averaged RoPE matrix over $T$ future positions.
\begin{lstlisting}[style=github, caption={Pytorch-like pseudo code for KV Cache compression with Expected Attention.}, label={lst:kv_compression}]
def compress(queries, keys, values, compression_ratio):
    # Compute query statistics
    mean_query, cov_query = compute_statistics(queries)    
    # Compute unnormalized attention scores (z_i)
    scores = matmul(mean_query, keys.T) / math.sqrt(d)
    scores += einsum("i,ij,j->", keys, cov_query, keys) / (2 * d)
    # Normalize scores and weight by value norms
    scores = softmax(scores, dim=-1) * values.norm(dim=-1)
    # Keep KV pairs with highest scores
    n_kept = int(keys.size(0) * (1 - compression_ratio))
    indices = scores.topk(n_kept, dim=-1).indices
    return keys[indices], values[indices]
\end{lstlisting}
\paragraph{Expected Attention Score}
With this query distribution, we can now analytically compute the expected unnormalized attention score in Equation~\ref{eq:attn_score}. For a query $\bar{q} \sim \mathcal{N}(\bar{\mu}_q, \bar{\Sigma}_q)$ in our interval $T$ and a fixed key $k_i$, the expected unnormalized score for that key is:
\begin{equation}
\label{eq:expected_z}
    \hat{z}_i = \mathbb{E}_{\bar{q} \sim \mathcal{N}(\bar{\mu}_q, \bar{\Sigma}_q)}\left[ \exp\left(\frac{\bar{q}^T k_i}{\sqrt{d}}\right) \right] = \exp\left(\frac{\bar{\mu}_q^T k_i}{\sqrt{d}} + \frac{k_i^T \bar{\Sigma}_q k_i}{2d}\right)
\end{equation}
%
%
where the second equality follows from the moment-generating function of a Gaussian distribution. 
We then define the expected attention score by applying the softmax on our unnormalized expectation:
\begin{equation}
\label{eq:expected_attention}
    \hat{a}_i = \frac{\hat{z}_i}{\sum_{j=1}^t \hat{z}_j}
\end{equation}
With this approximation, we can now estimate the importance of each cached KV pair. We define the expected contribution magnitude by substituting our expected attention weight into the contribution score formula from Equation~\ref{eq:score}:
\begin{equation}
\label{eq:contrib_score}
    \|\widehat{\Delta h}_i\| = (\hat{a}_i + \epsilon) \|W_o v_i\|
\end{equation}
where $\hat{a}_i$ is the expected attention weight from Equation~\ref{eq:expected_attention}, $\|W_o v_i\| \in \mathbb{R}$ is the magnitude of the transformed value vector, and $\epsilon$ is a small hyperparameter. This metric provides a tractable approximation to the true contribution score without requiring future queries.
\paragraph{Compression with Expected Attention} Equation~\ref{eq:contrib_score} captures the contribution of each KV pair to the transformer output. The Expected Attention compression algorithm scores all cached KV pairs according to Equation~\ref{eq:contrib_score} and evicts the $r \%$ pairs with the lowest expected contributions, where $r \in [0,1]$ is the compression ratio. Intuitively, this is equivalent to removing those KV pairs that have the smallest impact on the residual stream and therefore on the model output. We provide pseudo-code for our compression algorithm in \Cref{lst:kv_compression}.
%
%
%
%
%
%
%
%
\paragraph{Head-Adaptive Compression} Previous work has shown that different attention heads serve different roles in the model. We adopt adaptive per-layer compression~\citep{adakv} to account for this heterogeneity, allowing more important heads to retain more KV pairs.
\begin{figure}[t]
    \centering
    \begin{subfigure}{\linewidth}
        \includegraphics[width=\linewidth]{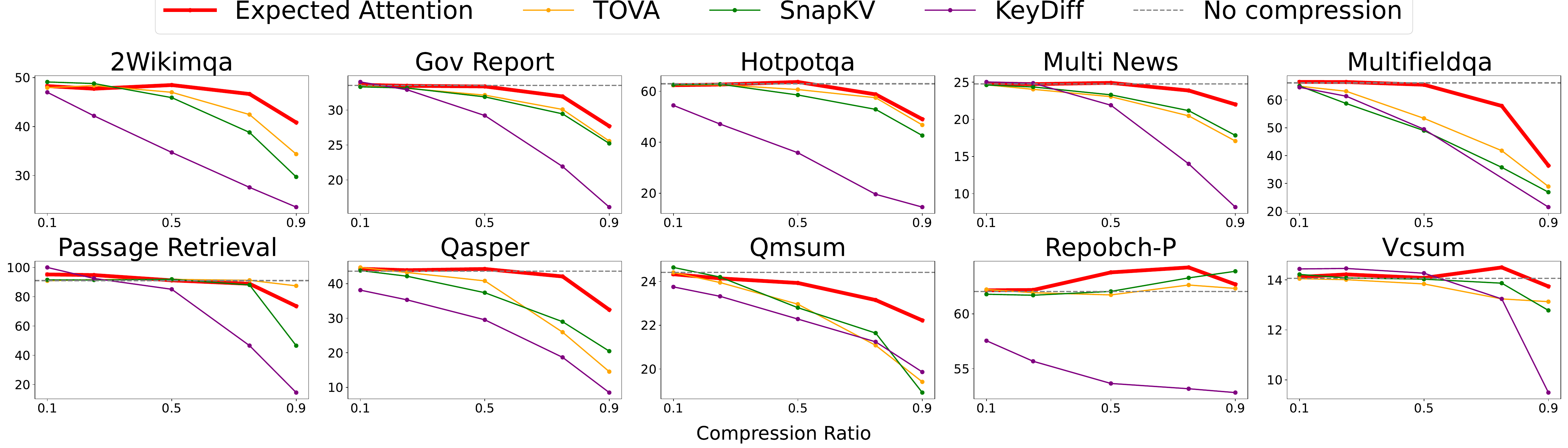}
    \end{subfigure}

    \vspace{-0.25cm} 

    \begin{subfigure}{\linewidth}
        \includegraphics[width=\linewidth]{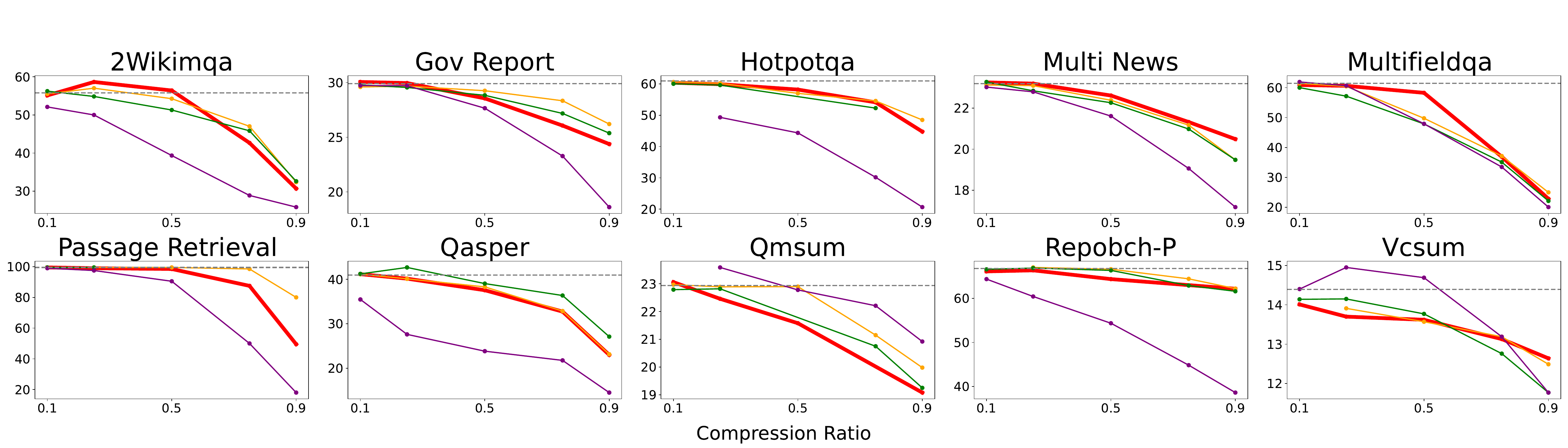}
    \end{subfigure}

    \caption{Scores on LongBench~\citep{longbench} for \qwen (top) and \gemma (bottom). The x-axis represents the compression ratio, the y-axis the score for each specific dataset. The horizontal line represents the baseline performance without cache compression. \ours achieves optimal trade-off between compression ratio and scores across most datasets (Additional and averaged results in \Cref{app:more_results}).}
    \label{fig:longbench}
\end{figure}
\section{Experiments}
\label{sec:experiments}
\subsection{Experimental Setup}
\paragraph{Prefilling vs Decoding Generation}
LLM inference comprises two phases with distinct computational characteristics. The \textit{prefilling phase} processes the entire input prompt in parallel, computing key-value projections for the \kvcache, a compute-bound operation requiring substantial floating-point operations. The \textit{decoding phase} sequentially generates tokens using the \kvcache and previous logits, appending new key-value pairs iteratively~\citep{howtoscale, vllminternals}. This dichotomy has motivated disaggregated architectures that implement prefill and decoding on different hardware, at the cost of transferring the cache, further incentivising compression~\citep{disaggregated, step3}. Therefore, an effective compression method must perform well in both prefilling and decoding~\citep{howtoscale, vllminternals}. Nevertheless, a number of recent methods often target a single phase: SnapKV~\citep{snapkv} for prefilling via query attention scores, StreamingLLM~\citep{streamingllm} and KNorm~\citep{knorm} for streaming decoding. \ours is designed considering these two aspects of LLM inference and addresses both scenarios efficiently. We present results for prefilling and decoding in \Cref{sec:pref} and \Cref{sec:dec} respectively.
\paragraph{Models and Datasets}
For prefilling (one-shot compression before generation), we test three model families supporting long contexts: \llama (128k)~\citep{llama3}, \qwen (32k)~\citep{qwen3}, and \gemma (128k)~\citep{gemma_2025}, all instruction-tuned. For decoding (compression during generation), we analyse reasoning models that generate extensive intermediate reasoning tokens and therefore large KV caches: \deeps, \deepsb~\citep{deepseekdistill}, and \omath~\citep{aimo}.

Our benchmarks include LongBench~\citep{longbench}, Ruler~\citep{ruler}, and Needle in a Haystack~\citep{niah, lostinthemiddle} for prefilling, and Aime25~\citep{balunovic_srimatharena_2025} and MATH-500~\citep{math500} for decoding.
\paragraph{Baselines}
Following an initial benchmarking study on Ruler (see \Cref{app:more_results}), we selected and compare our method against the best-performing baselines for each use case. For prefilling, we evaluate attention-based approaches like SnapKV~\citep{snapkv} and TOVA~\citep{tova}, embedding-based KeyDiff~\citep{keydiff}, and the trainable DuoAttention~\citep{duoattention} when the checkpoint is available.  {SnapKV}~\citep{snapkv} and {TOVA}~\citep{tova} rank KV pairs using attention scores from user queries. {KeyDiff}~\citep{keydiff} employs distance metrics between key embeddings for selection, making it also suitable for decoding generation. {DuoAttention}~\citep{duoattention} takes a trainable approach, learning compression masks for each attention head. For decoding, we focus on methods designed to be compatible with streaming generation: KNorm~\citep{knorm}, StreamingLLM~\citep{streamingllm}, and KeyDiff~\citep{keydiff}. {KNorm}~\citep{knorm} uses a simple approach by preserving keys with the lowest $L_2$ norm. {StreamingLLM}~\citep{streamingllm} maintains initial sink tokens throughout generation. 
\begin{table}[t]
\centering
\caption{\ours outperforms most baselines on Ruler~\citep{ruler} with 4K and 16K context length. We show average score with increasing compression ratios across baselines. Best results for each compression ratio are displayed in \textbf{bold}. The 0\% column indicates the baseline without compression.}
\label{tab:ruler}
\scriptsize 
\setlength{\tabcolsep}{4pt} 
\begin{tabularx}{\textwidth}{ll*{5}{>{\centering\arraybackslash}X@{\hspace{7mm}}}>{\centering\arraybackslash}X@{\hspace{1cm}}*{5}{>{\centering\arraybackslash}X@{\hspace{7mm}}}>{\centering\arraybackslash}X}
\toprule
\multirow{2}{*}{\textbf{Model}} & \multirow{2}{*}{\textbf{Method}} & \multicolumn{6}{c}{\textbf{Ruler 4k}} & \multicolumn{6}{c}{\textbf{Ruler 16k}} \\
\cmidrule(lr){3-14}
& & \textbf{0\%} & \textbf{10\%} & \textbf{25\%} & \textbf{50\%} & \textbf{75\%} & \textbf{90\%} & \textbf{0\%} & \textbf{10\%} & \textbf{25\%} & \textbf{50\%} & \textbf{75\%} & \textbf{90\%} \\
\toprule
    \multirow{4}{*}{\textit{\qwen}}  
    & EA (ours)     &  $\m{95.3}$   & $\m{95.3}  $ & $\m{95.0}  $ & $\m{94.7}  $ & $\m{88.3}  $ & $\m{65.4}  $    &  $\m{92.9}$   & $\m{93.1}  $ & $\m{93.2}  $ & $\m{92.7}  $ & $\m{85.6}  $ & $\m{62.7}  $ \\
    & TOVA[\citenum{tova}]  &  $\m{95.3}$     & $   89.0   $ & $   82.5   $ & $   77.6   $ & $   62.4   $ & $   24.7   $  &  $\m{92.9}$  & $   88.3   $ & $   81.7   $ & $   76.2   $ & $   68.7   $ & $   52.4   $ \\
    & SnapKV[\citenum{snapkv}]   &  $\m{95.3}$     & $   92.6   $ & $   84.0   $ & $   55.7   $ & $   33.1   $ & $   19.2   $   &  $\m{92.9}$  & $   90.1   $ & $   81.5   $ & $   62.8   $ & $   41.7   $ & $   26.8   $ \\
    & KeyDiff[\citenum{keydiff}]   &  $\m{95.3}$     & $   93.8   $ & $   89.4   $ & $   78.6   $ & $   64.4   $ & $   37.9   $   &  $\m{92.9}$  & $   88.9   $ & $   82.9   $ & $   74.5   $ & $   66.9   $ & $   53.1   $ \\    \midrule
    \multirow{4}{*}{\textit{\gemma}}
    &  EA (ours)      &  $\m{95.2}$ & $\m{95.2}  $ & $\m{94.9}  $ & $\m{92.7}  $ & $\m{78.2}  $ & $\m{53.6}  $  &  $\m{86.0}$ & $\m{82.8}  $ & $\m{81.7}  $ & $\m{76.6}  $ & $\m{60.5}  $ & $\m{41.8}  $ \\
    & TOVA[\citenum{tova}]    & $\m{95.2} $ & $   89.7   $ & $   81.1   $ & $   76.5   $ & $   58.1   $ & $   25.3   $  &  $\m{86.0}$ & $   79.7   $ & $   72.6   $ & $   62.5   $ & $   46.8   $ & $   32.7   $ \\
    & SnapKV[\citenum{snapkv}]   & $\m{95.2} $ & $   82.9   $ & $   72.0   $ & $   54.8   $ & $   40.3   $ & $   30.1   $   &  $\m{86.0}$ & $   74.1   $ & $   62.8   $ & $   46.4   $ & $   37.3   $ & $   31.4   $ \\
    & KeyDiff[\citenum{keydiff}]  & $\m{95.2} $ & $   94.3   $ & $   90.6   $ & $   79.8   $ & $   62.0   $ & $   34.3   $  &  $\m{86.0}$  & $   81.8   $ & $   78.6   $ & $   72.6   $ & $   58.6   $ & $   37.2   $ \\
    \midrule
    \multirow{5}{*}{\textit{\llama}} 
    &   EA (ours)      &  $\m{95.3}$   & $\m{95.7}  $ & $   95.3   $ & $   92.2   $ & $\m{75.9}  $ & $   30.6    $ &  $\m{93.4}$  & $\m{93.4}   $ & $   92.8   $ & $   86.0   $ & $   66.4    $ & $   25.5   $ \\
    & TOVA[\citenum{tova}]       &  $\m{95.3}$    & $   93.2   $ & $   87.3   $ & $   76.2   $ & $   63.3   $ & $   37.5   $ & $\m{93.4}$  & $   90.9    $ & $   86.1   $ & $   77.9   $ & $   68.4    $ & $   59.2   $ \\
    & Duo [\citenum{duoattention}]   &  $\m{95.3}$    & $   95.7   $ & $\m{95.7}  $ & $\m{95.3}  $ & $   73.2   $ & $   24.5   $ & $\m{93.4}$  & $   93.3    $ & $\m{93.0}  $ & $\m{90.1}  $ & $   59.1    $ & $   12.3   $ \\
    & SnapKV[\citenum{snapkv}]  &  $\m{95.3}$    & $   95.5   $ & $   88.8   $ & $   81.8   $ & $   63.2   $ & $   43.4   $ & $\m{93.4}$  & $   89.4    $ & $   82.0   $ & $   68.0   $ & $   43.1    $ & $   25.6   $ \\
    & KeyDiff[\citenum{keydiff}]  &  $\m{95.3}$    & $   94.7   $ & $   91.6   $ & $   85.5   $ & $   72.9   $ & $\m{61.1}  $ & $\m{93.4}$  & $   92.1    $ & $   88.4   $ & $   82.6   $ & $\m{74.9}   $ & $\m{66.5}  $ \\
 \bottomrule
\end{tabularx}
\end{table}
\paragraph{Implementation details}
We implement \ours in Pytorch~\citep{pytorch}. For all benchmarks, we test the models on 8 H100 GPUs, with batch size 1. We make all the code to reproduce our method and the baselines available in KVPress. In all experiments we use $\epsilon = 0.02$, except for needle in a haystack where use $\epsilon=0$, and we average the RoPE embeddings over the next $T=512$ positions. For prefilling, we do not assume any question about the context. This simulates a real world use case and avoids favouring methods like SnapKV that rely on this assumption. For decoding, we keep a small buffer of hidden states of 128 tokens to compute statistics, and perform compression every 512 generation steps.  In Equation~\ref{eq:contrib_score} we only use $V$ instead of $W_o V$, as using $W_o$ leads to a minor increase in results at a significantly higher memory cost.

\section{Experimental Results}
\subsection{Prefilling}
\label{sec:pref}
\begin{figure}[t]
    \centering
    \begin{subfigure}[b]{0.255\textwidth}
        \centering
        \includegraphics[width=\textwidth]{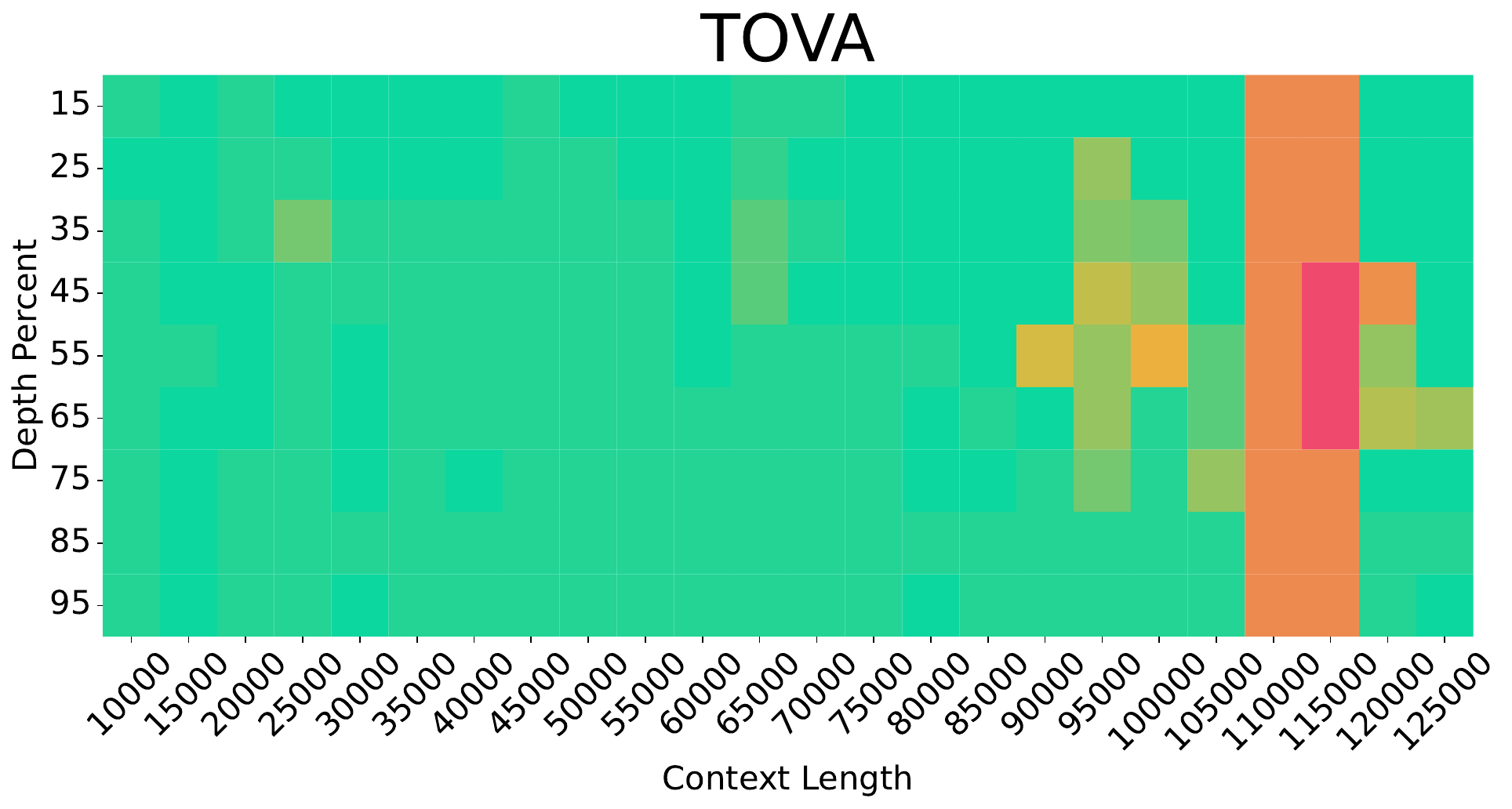}
    \end{subfigure}
    \begin{subfigure}[b]{0.24\textwidth}
        \centering
        \includegraphics[width=\textwidth]{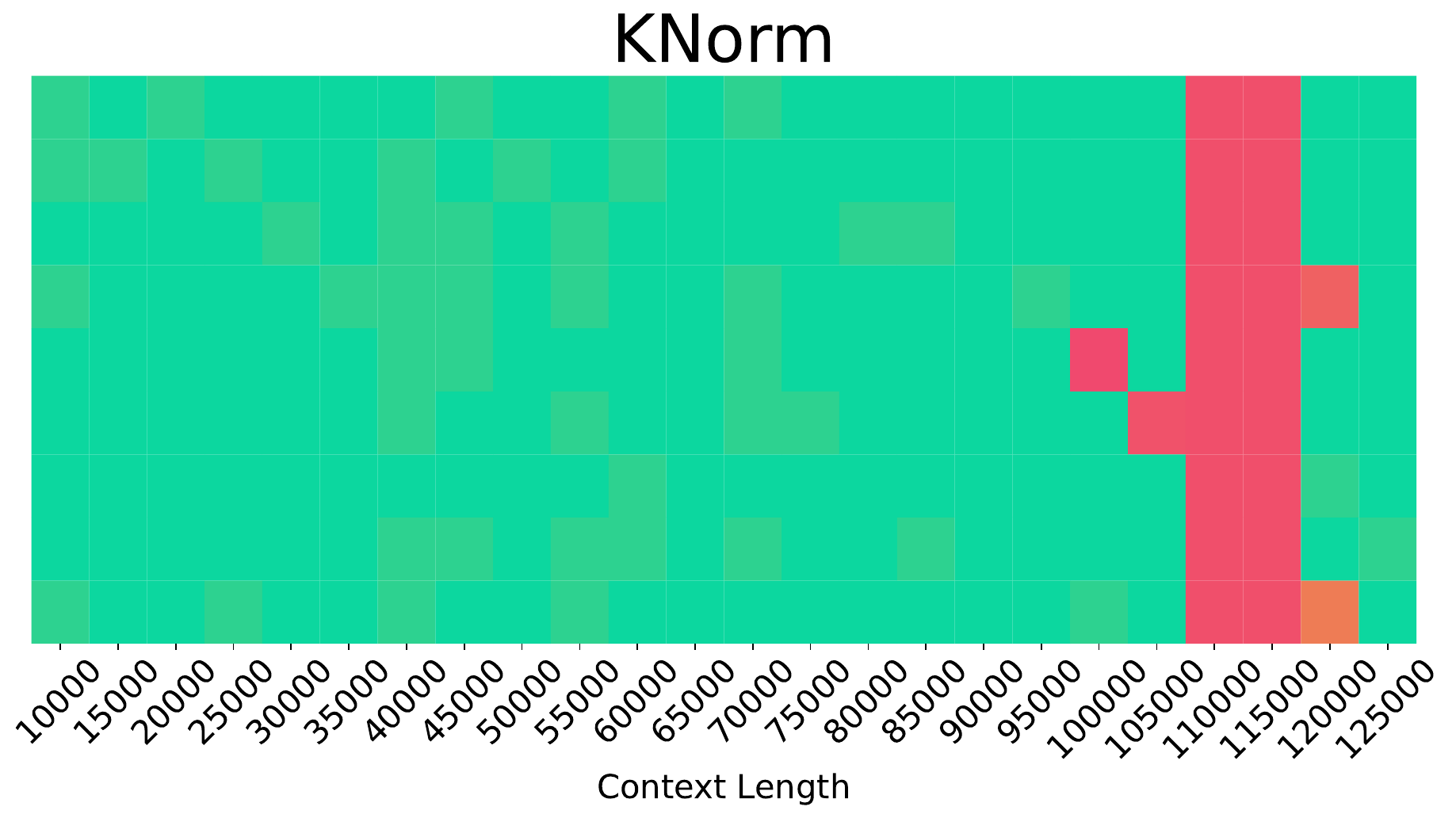}
    \end{subfigure}
    \begin{subfigure}[b]{0.24\textwidth}
        \centering
        \includegraphics[width=\textwidth]{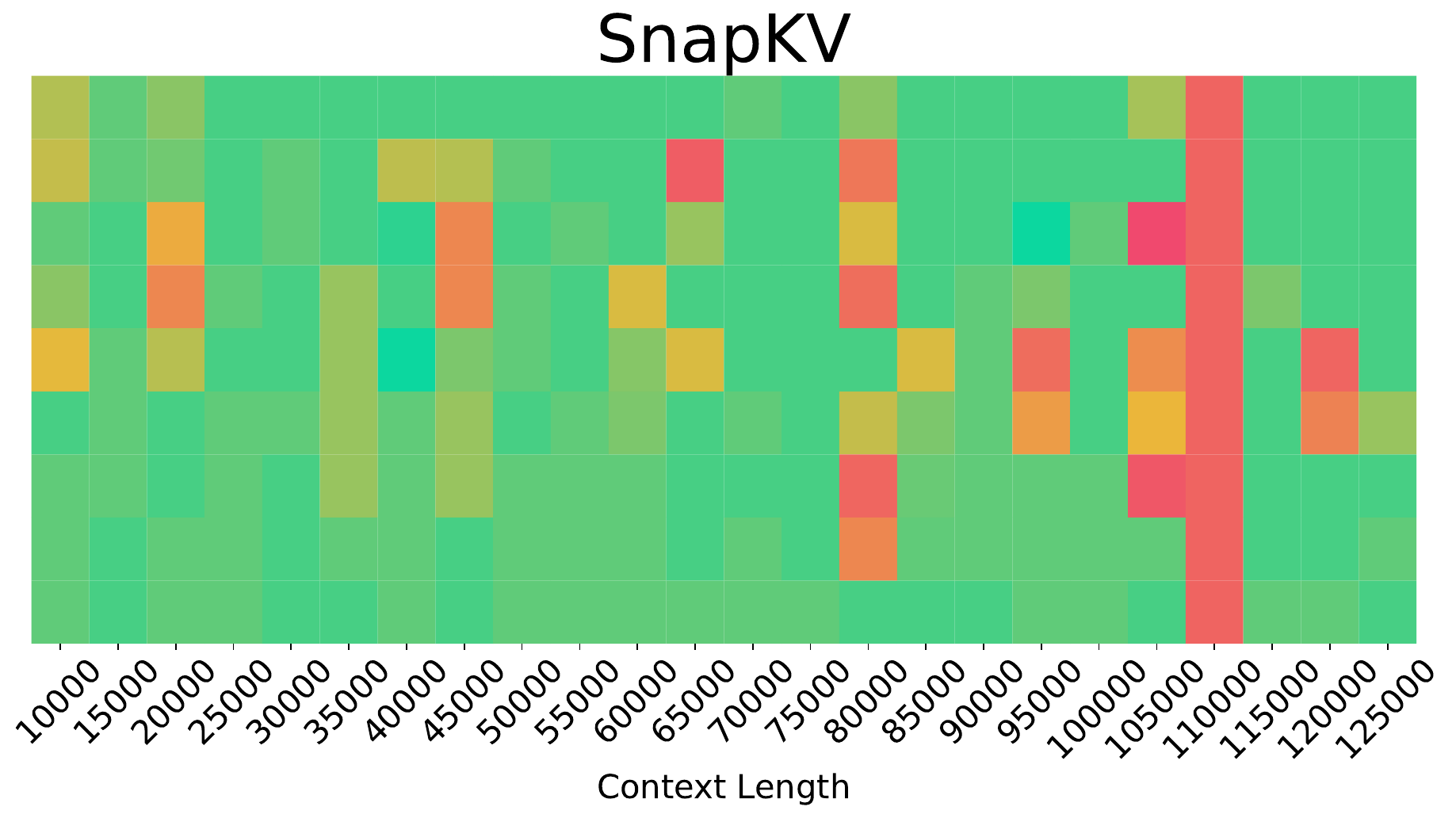}
    \end{subfigure}
    \begin{subfigure}[b]{0.24\textwidth}
        \centering
        \includegraphics[width=\textwidth]{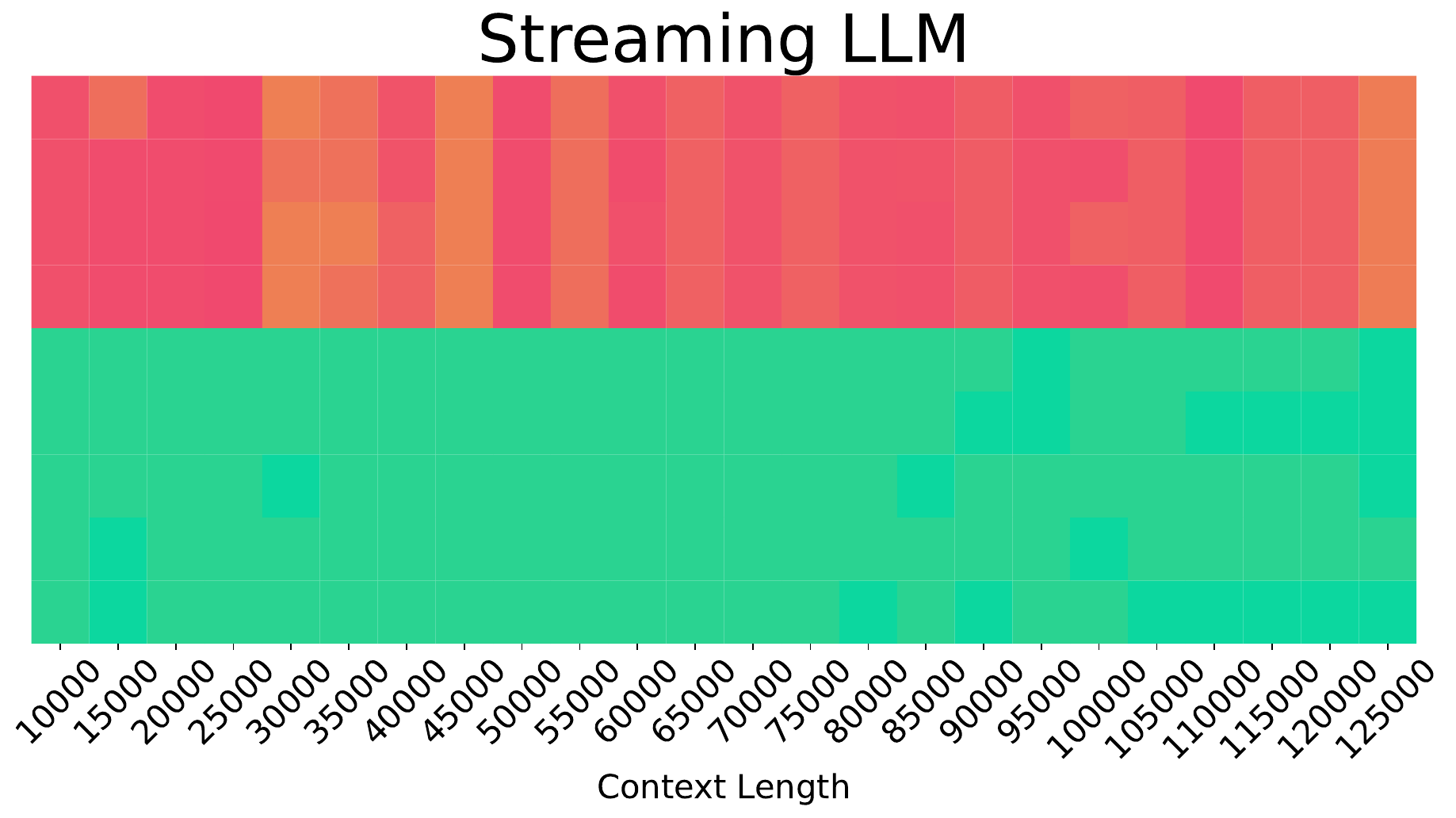}
    \end{subfigure}
    
    \begin{subfigure}[b]{0.255\textwidth}
        \centering
        \includegraphics[width=\textwidth]{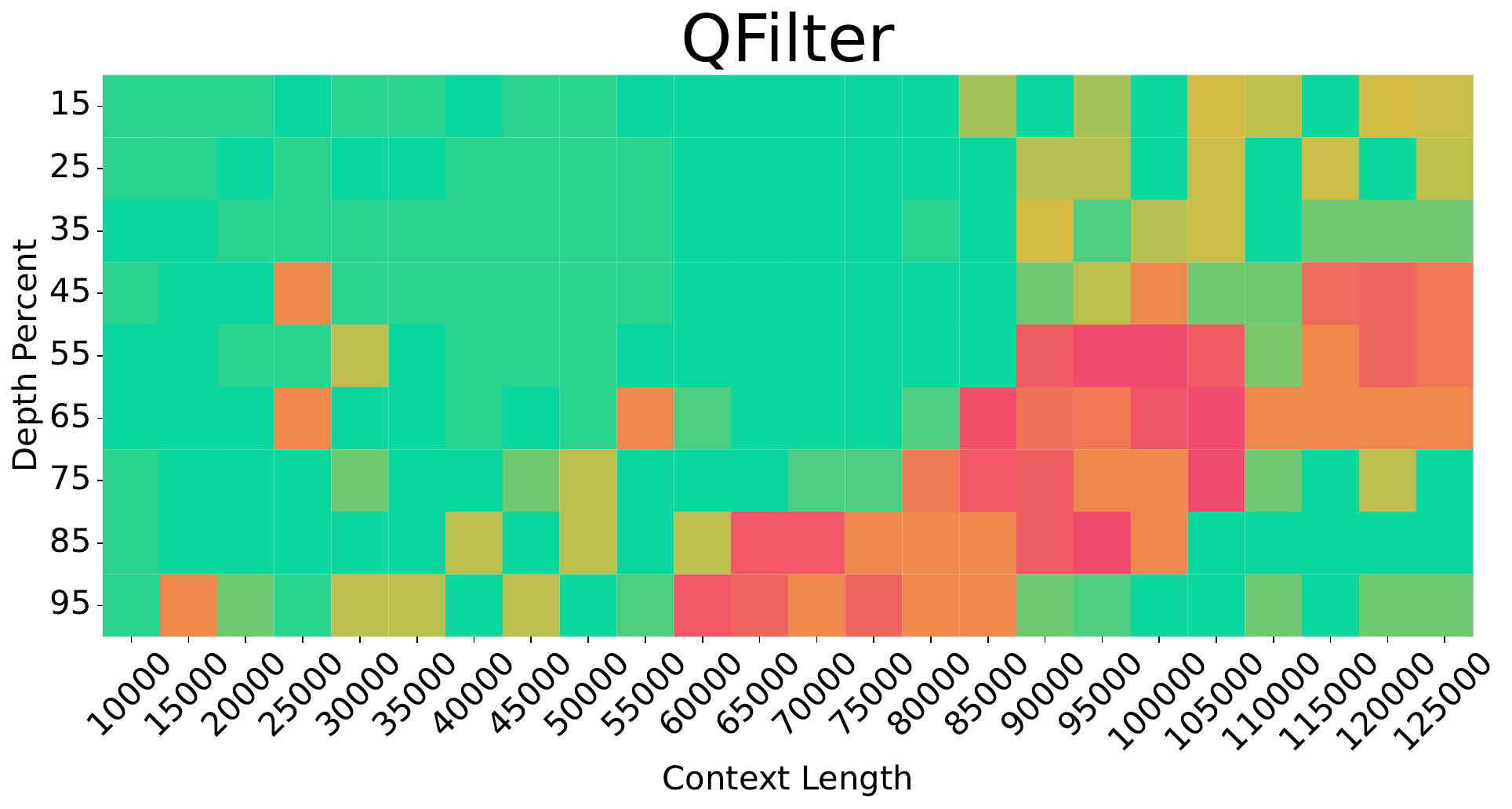}
    \end{subfigure}
    \begin{subfigure}[b]{0.24\textwidth}
        \centering
        \includegraphics[width=\textwidth]{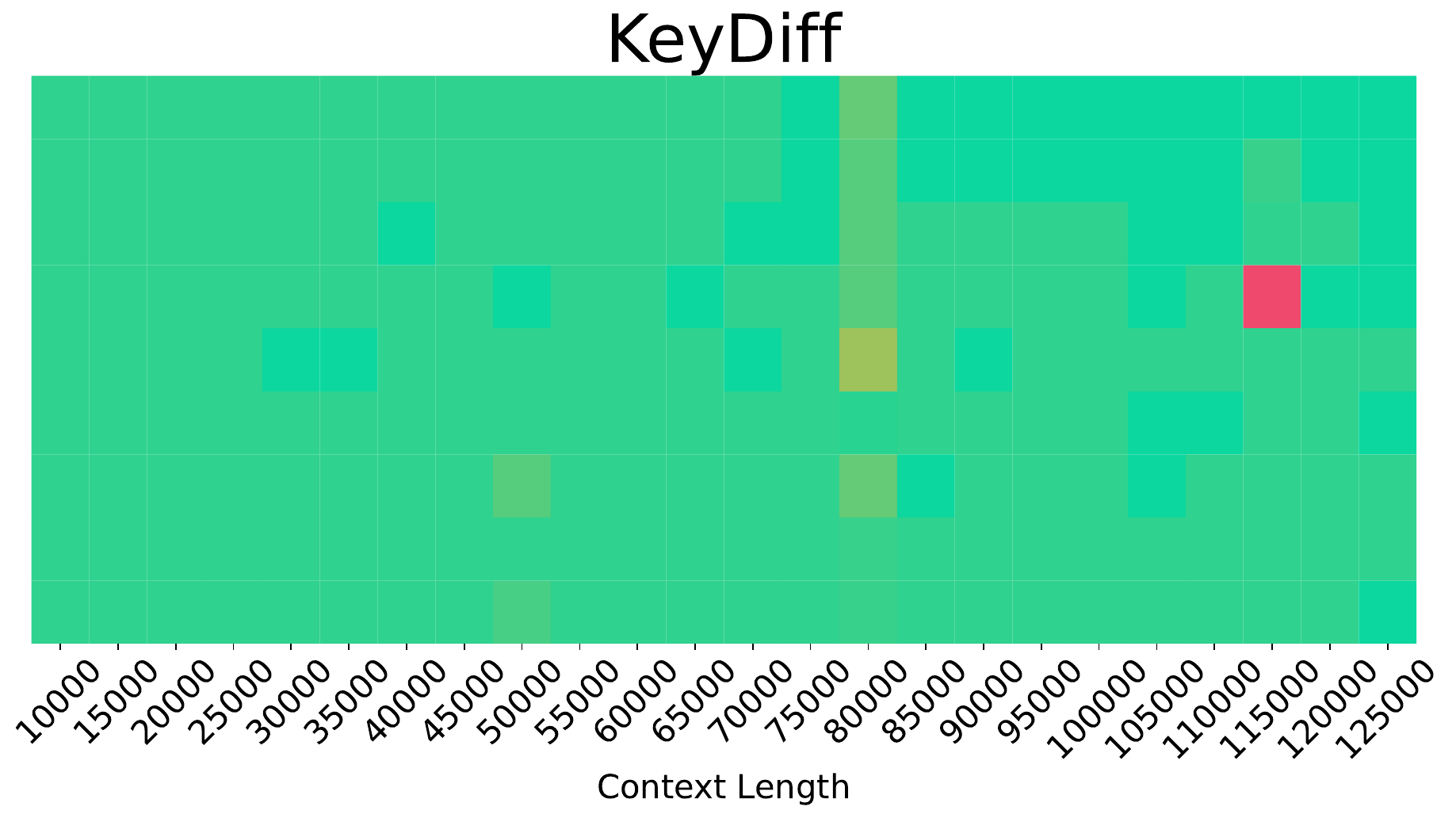}
    \end{subfigure}
    \begin{subfigure}[b]{0.24\textwidth}
        \centering
        \includegraphics[width=\textwidth]{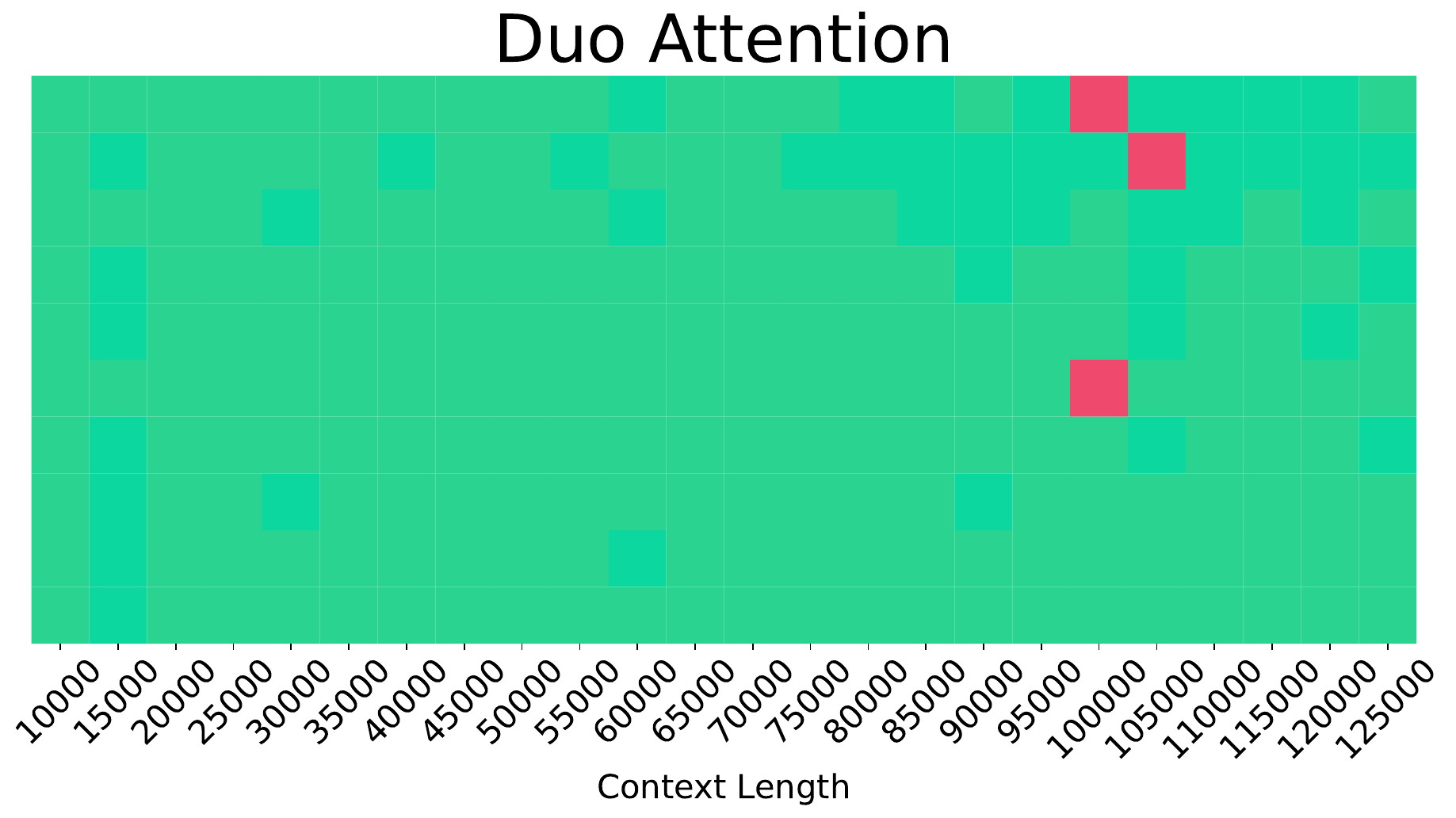}
    \end{subfigure}
    \begin{subfigure}[b]{0.24\textwidth}
        \centering
        \includegraphics[width=\textwidth]{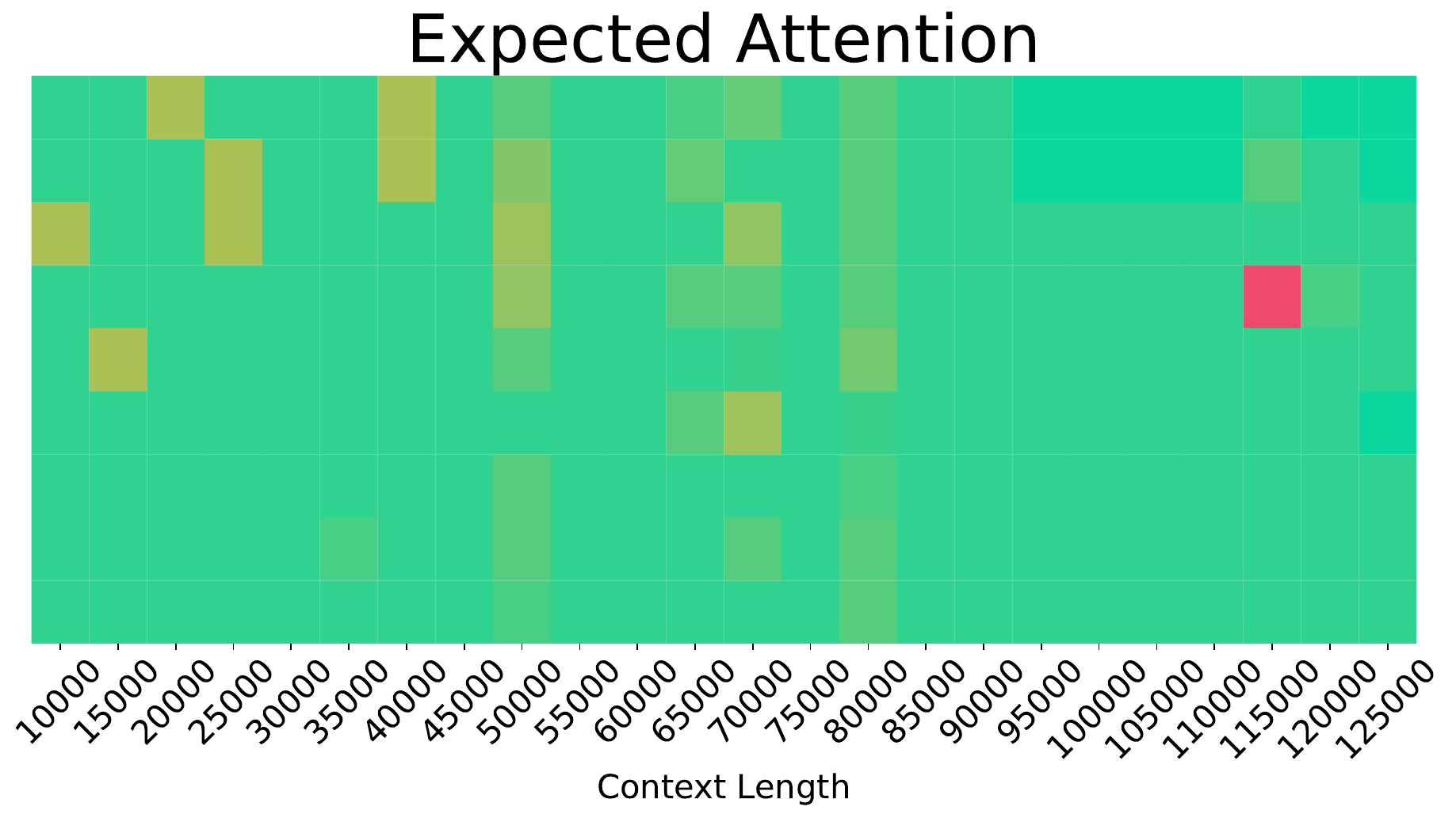}
    \end{subfigure}
    
    \caption{Needle in the Haystack test for different methods with \llama and 50\% compression ratio.}
    \label{fig:llama_niah}
\end{figure}

\paragraph{LongBench}
LongBench~\citep{longbench} tests long-context capabilities across diverse tasks. The benchmark comprises six categories: single and multi-document QA, summarization, few-shot learning, synthetic tasks, and code completion. As shown in \Cref{fig:longbench} for \llama and \qwen (see \Cref{app:more_results} for \gemma), \ours consistently achieves optimal compression-performance trade-offs, maintaining higher scores across all compression ratios. This demonstrates effective retention of critical KV pairs even under significant compression across varied reasoning and generation tasks.
\paragraph{Ruler}
Ruler~\citep{ruler} measures retrieval, multi-hop tracing, and aggregation abilities within long contexts through four subsets: NIAH (Needle-in-a-Haystack) for single-fact retrieval, VT (Variable Tracking) for multi-hop reasoning, CWE (Common Words Extraction) for frequency-based aggregation, and FWE (Frequent Words Extraction) for statistical pattern recognition. \Cref{tab:ruler} shows results at various compression ratios for 4k and 16k windows. \ours maintains strong performance across all subsets, particularly at higher compression ratios. While KeyDiff performs well on \llama, it struggles on \gemma and \qwen, potentially due to QK normalization~\citep{gemma_2025, qwen3}. Our Expected Attention-based policy effectively preserves information necessary for precise retrieval tasks.
\paragraph{Needle in a Haystack} 
The NIAH test~\citep{niah} embeds specific information (the "needle") within lengthy distracting text (the "haystack") to evaluate retrieval capabilities across varying context positions and lengths. The test systematically varies both the needle's position within the context (needle depth) and the total context length to assess consistent retrieval performance. \Cref{fig:llama_niah} visualizes retrieval success across needle positions and context lengths up to 125k tokens. \ours demonstrates robust performance comparable to DuoAttention and significantly more stable than other baselines in long contexts, confirming retention of critical information under compression regardless of needle placement or context size.
\begin{figure}[t]
    \begin{minipage}[c]{0.51\textwidth}
        \centering
        \captionof{figure}{Decoding results on Aime25 dataset, different markers represent different models sizes. The x-axis is the maximum size that the \kvcache is allowed to grow to.}
        \hspace{-1em}
        \includegraphics[width=\textwidth]{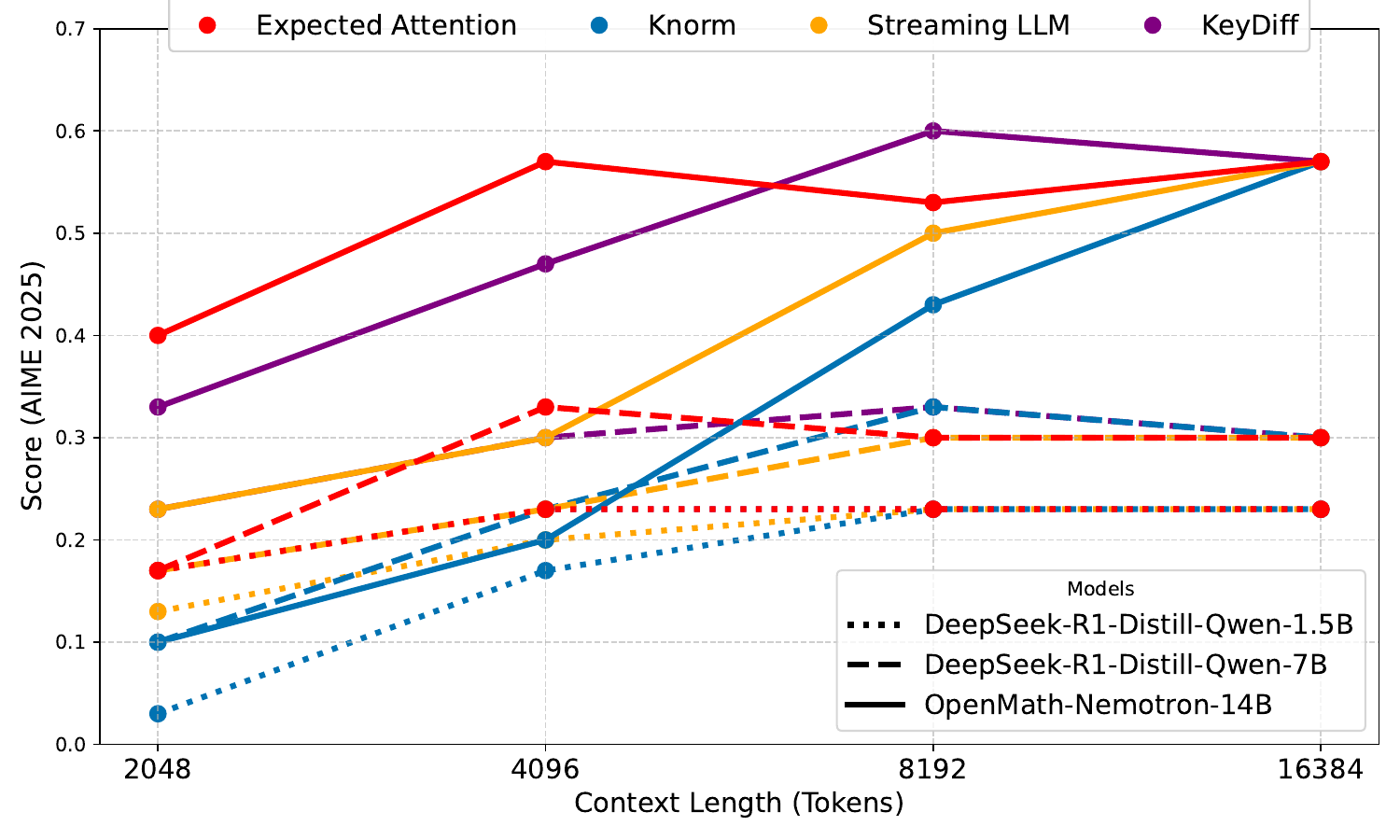}
        \label{fig:decoding}
    \end{minipage}
    \hspace{0.5em}
    \begin{minipage}[c]{0.44\textwidth}
        \centering
        \captionof{table}{Decoding scores on MATH-500. Columns indicate the final size of the \kvcache with respect to the original full version. Best scores in \textbf{bold}.}
        \label{tab:decoding}
        \scalebox{0.7}{
        \begin{tabular}{ll|c|ccc}
            \toprule
            \textbf{Model} & \textbf{Method} & \multicolumn{4}{c}{\textbf{Compression}} \\
            & & \textbf{0$\times$}  & \textbf{2$\times$} & \textbf{4$\times$} & \textbf{12$\times$} \\
            \midrule
            \multirow{4}{*}{Qwen-R1-1.5B}
            &  \cellcolor{gray!25}  EA (ours) & \cellcolor{gray!25}  \textbf{0.47} & \cellcolor{gray!25}  $\m{0.47}$ & \cellcolor{gray!25}  $\m{0.43}$ & \cellcolor{gray!25} $\m{0.33}$ \\
            & KeyDiff[\citenum{keydiff}] & \textbf{0.47} & 0.42 & 0.40 & 0.30 \\
            & KNorm[\citenum{knorm}] & \textbf{0.47} & 0.41 & 0.28 & 0.11 \\
            & Streaming[\citenum{streamingllm}] & \textbf{0.47} & 0.45 & 0.41 & 0.31 \\
            \midrule
            \multirow{4}{*}{Qwen-R1-7B}
            & \cellcolor{gray!25}  EA (ours) & \cellcolor{gray!25}  \textbf{0.57} & \cellcolor{gray!25}  $\m{0.55}$ & \cellcolor{gray!25}  $\m{0.53}$ & \cellcolor{gray!25} $\m{0.49}$ \\
            & KeyDiff[\citenum{keydiff}] & \textbf{0.57} & 0.54 & 0.48 & 0.35 \\
            & KNorm[\citenum{knorm}] & \textbf{0.57} & 0.47 & 0.32 & 0.12 \\
            & Streaming[\citenum{streamingllm}] & \textbf{0.57} & 0.54 & 0.51 & 0.41 \\
            \midrule
            \multirow{4}{*}{Nemotron-14B}
            & \cellcolor{gray!25}  EA (ours) & \cellcolor{gray!25}  \textbf{0.57} & \cellcolor{gray!25} 0.55 & \cellcolor{gray!25}  $  \m{0.54}$ & \cellcolor{gray!25} $\m{0.47}$ \\
            & KeyDiff[\citenum{keydiff}] & \textbf{0.57} & 0.56 & 0.51 & 0.44 \\
            & KNorm[\citenum{knorm}] & \textbf{0.57} & 0.50 & 0.36 & 0.14 \\
            & Streaming[\citenum{streamingllm}] & \textbf{0.57} & $\m{0.57}$ & $\m{0.54}$ & 0.42 \\
            \bottomrule
        \end{tabular}}
        \end{minipage}
\end{figure}
\subsection{Decoding}
\label{sec:dec}
For decoding, we evaluate \ours on reasoning models, \deeps, \deepsb, and \omath. Reasoning models are particularly suitable for our evaluation as they generate extensive chain-of-thought outputs for reasoning traces, placing significant demands on \kvcache memory~\citep{dms}. We use the Aime25~\citep{yamada2025aiscientistv2workshoplevelautomated} and MATH-500~\citep{math500} datasets. Aime25 consists of competition-level mathematical problems requiring multi-step reasoning and precise calculation, while MATH-500 encompasses diverse mathematical domains with varying difficulty levels. During decoding, we allow the \kvcache to expand to a predetermined size before initiating token eviction. In the tables, we use $n\times$ to show that the final cache size is $n$ times smaller than would be without compression.

Results for Aime25 and MATH-500 are presented in \Cref{fig:decoding} and \Cref{tab:decoding}, respectively. \ours consistently outperforms or matches baseline methods across all models, with particularly strong performance at higher compression ratios ($4\times$ and $16\times$). Most methods demonstrate minimal performance degradation at $2\times$ compression, indicating that a large portion of tokens in reasoning traces contains redundant information that can be pruned without affecting mathematical reasoning performance. \ours shows the best performance especially in high-compression scenarios (12$\times$ compression).
\subsection{Memory Savings and Efficiency}
We evaluate the memory efficiency of our method using \llama and \qwen for both prefilling and decoding phases. All experiments are conducted on a single H100 GPU with bfloat16 precision for both model weights and \kvcache. We focus on peak memory usage as the primary efficiency metric, as \kvcache memory consumption is often the primary bottleneck for long-context inference.

\begin{figure}[t]
    \centering
    \subcaptionbox{Peak memory usage vs sequence length up to 120k for \llama, with 50\% and 90\% compression ratio. As the context length grows the memory savings become more evident, achieving up to 15GB less memory for large contexts. \label{fig:peak}}[0.48\linewidth]
        {\includegraphics[width=0.48\textwidth]{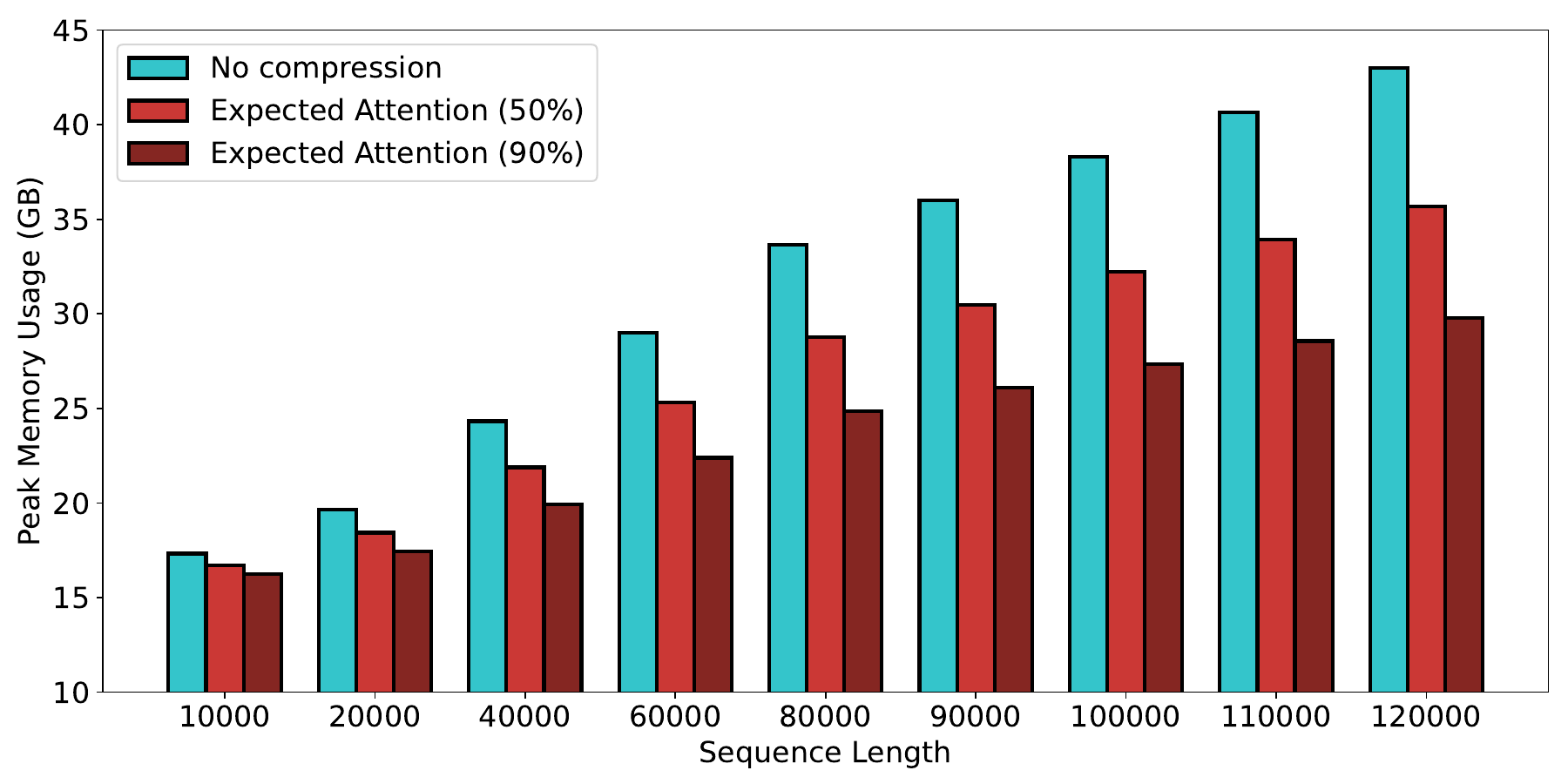}}
    \hfill 
    \subcaptionbox{Needle in a Haystack score with different compression ratios with \qwen. \ours has no accuracy loss with a compression ratio of 50\%. Marker size indicates actual \kvcache size in GB. \label{fig:niaheff}}[0.49\linewidth]
        {\includegraphics[width=0.49\textwidth]{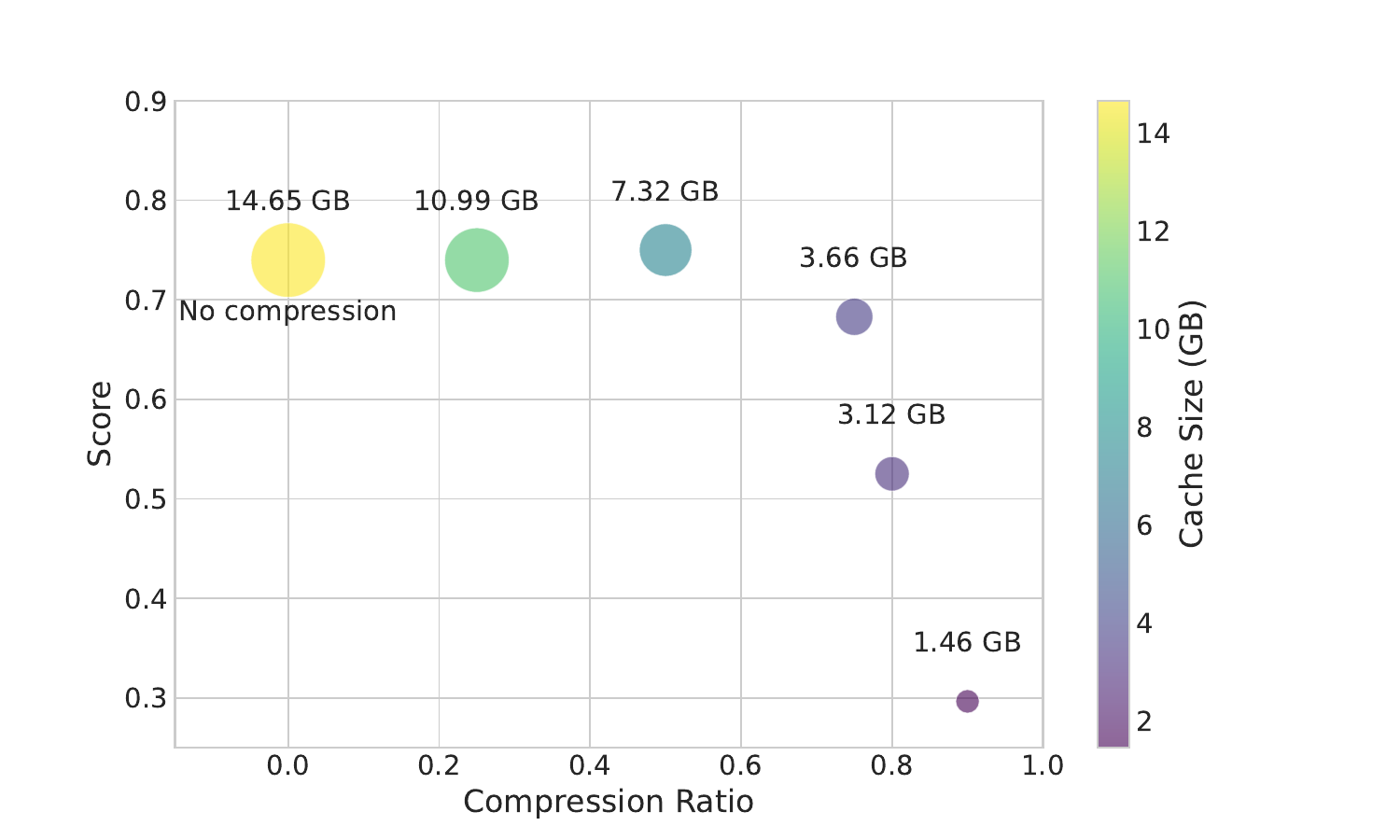}}
    \label{fig:efficiency}
\end{figure}

\Cref{fig:peak} demonstrates peak memory usage as sequence length increases up to 120k tokens, comparing \ours at 50\% and 90\% compression ratios against the uncompressed baseline with vanilla attention. The results show that memory savings become increasingly substantial as context length grows.

\Cref{fig:niaheff} illustrates the relationship between compression ratio (x-axis) and NIAH benchmark performance for \qwen, with marker size representing the corresponding \kvcache size. While higher compression ratios naturally reduce \kvcache size, they typically incur performance penalties. Remarkably, \ours at 50\% compression maintains performance parity with the uncompressed baseline while achieving a $2\times$ reduction in \kvcache size, demonstrating an optimal balance between memory efficiency and task performance.

\section{KVPress: A Research Framework for KV Cache Compression}
We introduce \href{https://github.com/NVIDIA/kvpress}{\textbf{KVPress}}, a comprehensive PyTorch-based library designed to streamline the development and benchmarking of \kvcache compression methods. By natively integrating with Hugging Face transformers~\citep{transformers}, KVPress allows researchers to implement and test novel techniques rapidly and intuitively.

KVPress achieves this integration by utilizing PyTorch forward hooks attached to each attention layer. This design choice allows the framework to operate entirely within the existing Hugging Face transformer pipeline, eliminating the need to modify the model architecture or implement a custom, low-level KV cache management system. Specifically, after each attention layer's forward pass, the hook is triggered to calculate the importance scores for the current KV pairs. Based on a chosen compression policy (e.g., \ours), the hook then selectively evicts the KV pairs with the lowest scores before the model proceeds to the next layer. This non-invasive, layer-by-layer compression mechanism significantly simplifies the experimentation process for researchers. KVPress currently incorporates over 20 existing state-of-the-art compression techniques, including post-training and trainable ones.
Our primary goal is to establish a shared, standardized platform that accelerates research and ensures fair, reproducible benchmarking within the $\mathrm{KV}$ Cache compression field.

It is important to note that KVPress is not an optimized engine for deployment. This choice to prioritize readability over runtime efficiency facilitates the design of standard implementations and consistent benchmarking. We believe production-level efficiency is best achieved through subsequent, low-level optimization of individual methods, which would otherwise complicate the framework's primary role as a unified research tool.
\subsection{KVPress Leaderboard and Standardized Benchmarking}
To complement the framework, we provide a public \href{https://huggingface.co/spaces/nvidia/kvpress-leaderboard}{\textbf{KVPress leaderboard}} for standardized evaluation across multiple long-context benchmarks. The leaderboard establishes consistent evaluation protocols, enabling direct and reproducible comparison of new compression methods against existing approaches. We hope that this framework, alongside its standardized benchmarking suite, will support the research community's efforts to develop and validate novel compression techniques for long-context language models.
\section{Related Works}
\paragraph{Trainable KV-Cache Compression}
One approach to reducing memory requirements involves modifying the model architecture or training procedure to inherently produce smaller caches. \citet{mqa, mmqa} reduce cache size by decreasing the number of key-value heads, effectively sharing key-value representations across queries. DeepSeek-V2~\citep{deepseekv3} introduced Multi-Head Latent Attention, which projects keys and values into a lower-dimensional latent space during training, directly reducing the memory footprint of cached representations. Alternative trainable approaches focus on learning compression policies~\citep{dms, dmc} or masks~\citep{duoattention}  from pre-trained checkpoints. Finally, State Space Models~\citep{ssm,mamba} replace the quadratic attention mechanism with linear-complexity alternatives, while hybrid approaches combine transformer layers with RNN-based components~\citep{phi4flash, zamba}. Although these trainable methods typically achieve superior performance, they require substantial computational resources for pre-training or continued pre-training, making them less practical for deployment with existing large-scale models.
\paragraph{Training-Free \kvcache compression}
Given the computational costs associated with trainable methods, significant research effort has focused on developing post-training compression techniques that can be applied to existing models without modification. Early approaches~\citep{snapkv, tova} directly utilize attention scores to rank KV pairs by importance. However, these methods require access to the full attention matrix, making them incompatible with Flash Attention~\citep{flashattention} and thus impractical for modern deployment scenarios. To address this limitation, several works have developed heuristic-based importance measures that can be computed without materializing attention matrices, such as keys norm (KNorm~\citet{knorm}), token positions (StreamingLLM~\citet{streamingllm}, H2O~\citet{h2o}) or SVD projection (Q-Filters~\citet{qfilters}). Recognizing that different attention heads exhibit varying sensitivity to compression, recent methods such as AdaKV~\citep{adakv} and PyramidKV~\citep{pyramidkv}  adopt head-specific compression strategies. \textit{\ours}, adopts insights from these heuristic approaches while providing a principled theoretical foundation based on the distributional properties of transformer activations.
\paragraph{Quantization}
Instead of reducing the \kvcache size along the sequence dimension, quantization methods try to reduce the precision used to store the cache. For example, NQKV \citet{nqkv} partitions the cache into blocks for quantization and processes them separately. KVQuant~\citep{kvquant} performs non uniform per-layer quantization, while KIVI~\citep{kivi} quantizes the key cache by layer and the value cache by token. These methods are orthogonal to \ours (and to \kvcache compression in general), making it possible to integrate them. 
\paragraph{Efficient Implementations}
Alongside compression, sparse attention and quantization, another effort has been done to devise efficient implementation of inference systems. In this context, a well designed low-level handling of the \kvcache can deliver significant performance speed-ups, especially in multi-user serving systems. The first to investigate this and introduce efficient memory management for \kvcache was vLLM~\citep{vllm}, soon followed by other approaches~\citep{vattention, minference} and frameworks~\citep{trtllm}.
\section{Limitations}
A key trade-off of our training-free methodology is that its performance does not match that of trainable methods~\citep{deepseekv2, dms}. This is an intentional design choice that allows deployment without significant computational resources required for intensive training. Future work could explore combining our theoretical framework with lightweight fine-tuning.

Another limitation is that our method requires users to specify compression ratios manually, lacking an automated mechanism to determine optimal compression levels for different scenarios such as text generation. This represents a promising area for future research.

Finally, while our PyTorch implementation effectively demonstrates our method's theoretical principles, it is not optimized for efficiency. A highly performant implementation with custom CUDA kernels would significantly improve speed and practical utility.
\section{Conclusion}
We introduced \ours, a training-free algorithm for \kvcache compression. We showed \ours outperforms state-of-art \kvcache compression methods on several benchmarks and in both prefilling and decoding scenarios. Additionally, we released a research library that allows researchers to easily implement and experiment with \kvcache compression methods, and evaluate them on popular benchmarks for long context.
\bibliographystyle{style}
\bibliography{references}

\clearpage
\appendix
\section{Reconstruction Error Across Methods}
\label{app:approximations}
\begin{figure}[t]
    \centering
    \begin{subfigure}[b]{0.32\linewidth}
        \centering
        \includegraphics[width=\linewidth]{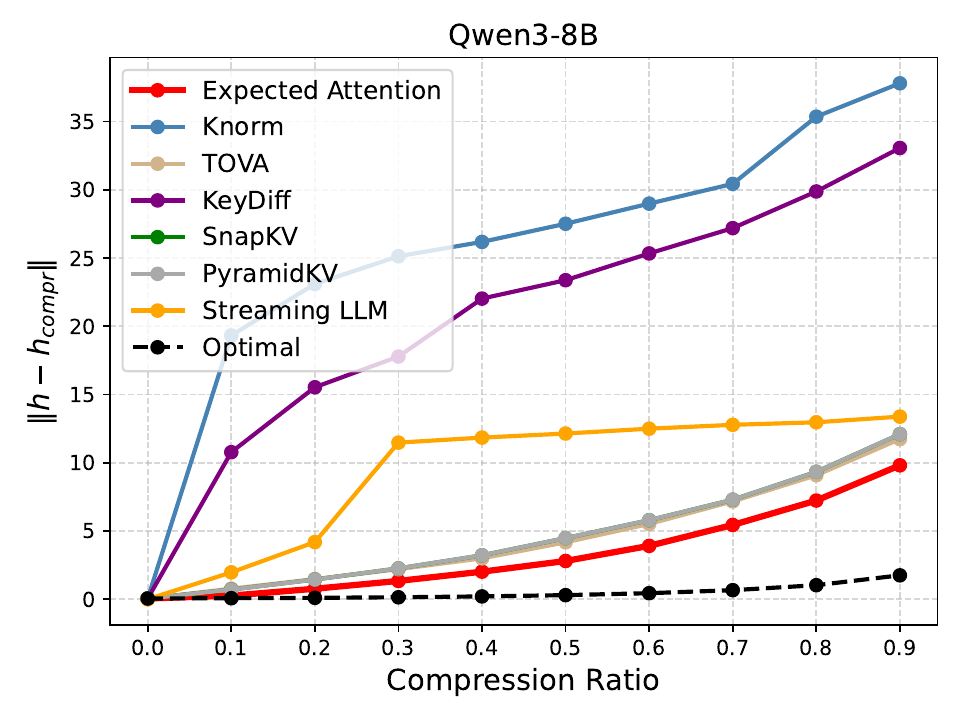}
    \end{subfigure}
    \begin{subfigure}[b]{0.32\linewidth}
        \centering
        \includegraphics[width=\linewidth]{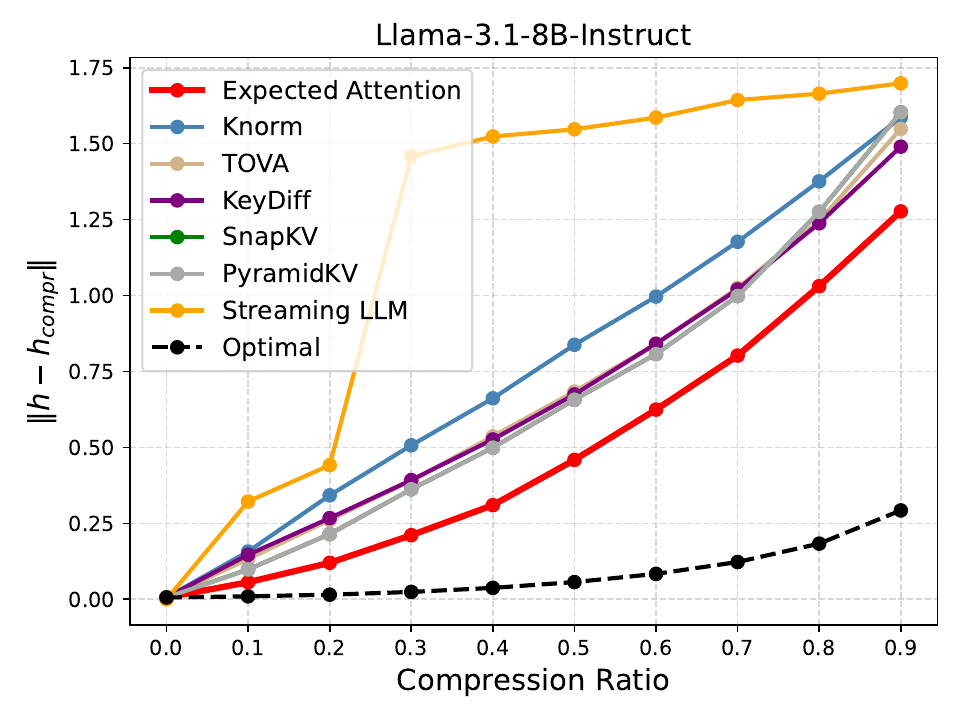}
    \end{subfigure}
    \begin{subfigure}[b]{0.32\linewidth}
        \centering
        \includegraphics[width=\linewidth]{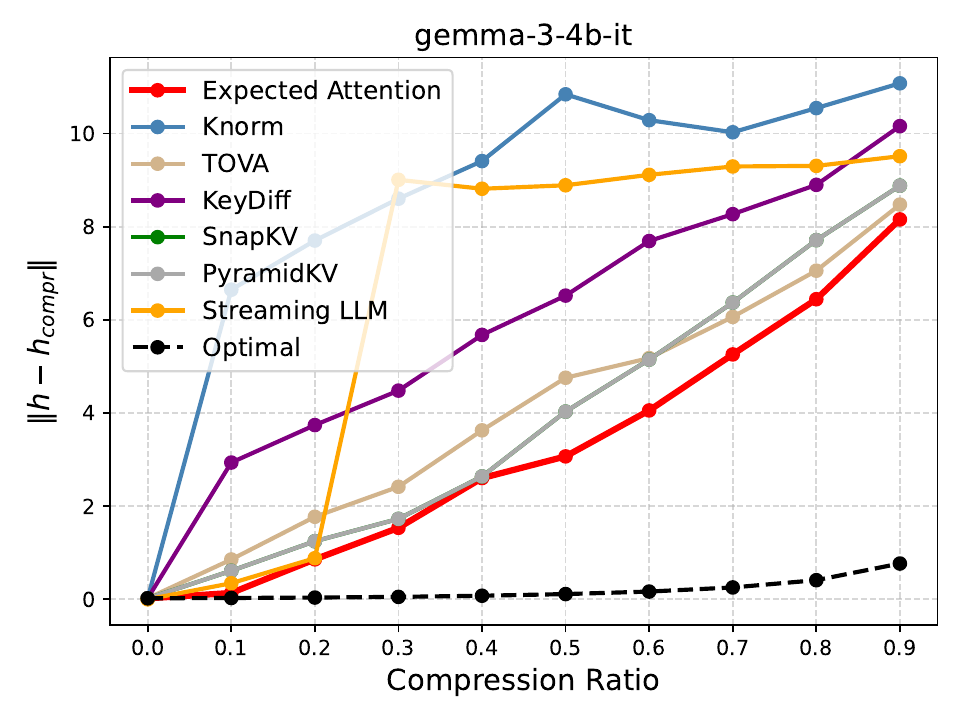}
    \end{subfigure}
\caption{Reconstruction error $\| h - h_{\text{compr}} \|$} averaged across model layers. \ours achieves the best error, minimizing the impact on the residual stream.
\label{fig:reconstr_err}
\end{figure}

In \Cref{sec:expected_attention}, we discussed the challenge of compressing the \kvcache without significantly altering the residual stream. To understand the impact of \ours on the model output, we quantify the reconstruction error of the residual stream, i.e. how the difference between the original, uncompressed hidden states and the corresponding hidden states after compression. We define the reconstruction error as $\| h - h_{\text{compr}} \|$, where $h$ is the original hidden state without compression and $h_{\text{compr}}$ the hidden state after the \kvcache has been compressed. We average the reconstrcution error over a long sequence of $\sim$ 5K tokens and display the results for several methods in \Cref{fig:reconstr_err}. \ours consistently achieves a lower reconstruction error, indicating that it preserves the integrity of the hidden state more effectively than competing methods, a crucial property for maintaining downstream performance \citep{limitations, vllminternals}.
\section{Distributional Properties of LLM activations}
\label{app:adistributions}
In this section, we analyse the distributional properties of activations within Large Language Models. Our investigation aligns with the findings of prior work, which has demonstrated that LLM activations often exhibit normal distributions. More specifically \citet{teal} finds that hidden states are zero-mean unimodal, and qualitatively fall into two distinctly shaped distributions. The hidden states before the Attention and the MLP layers tend to be Gaussian-like, while the hidden states in the intermediate of such layers tend to be Laplacian-like.

\begin{figure}[t]
    \centering
    \begin{subfigure}[t]{\linewidth}
        \centering
        \includegraphics[width=\linewidth]{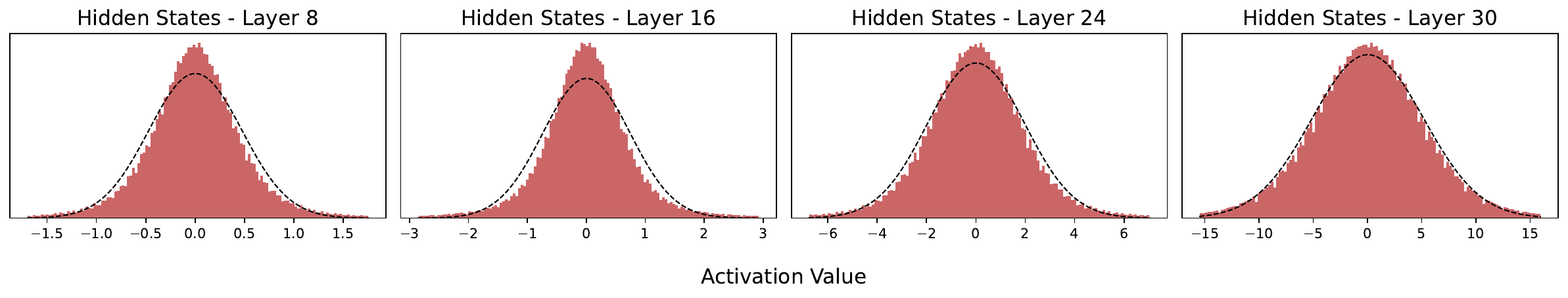}
        \caption{\qwen Hidden States distributions.}
        \label{fig:qwen_hs}
    \end{subfigure}
    \begin{subfigure}[t]{\linewidth}
        \centering
        \includegraphics[width=\linewidth]{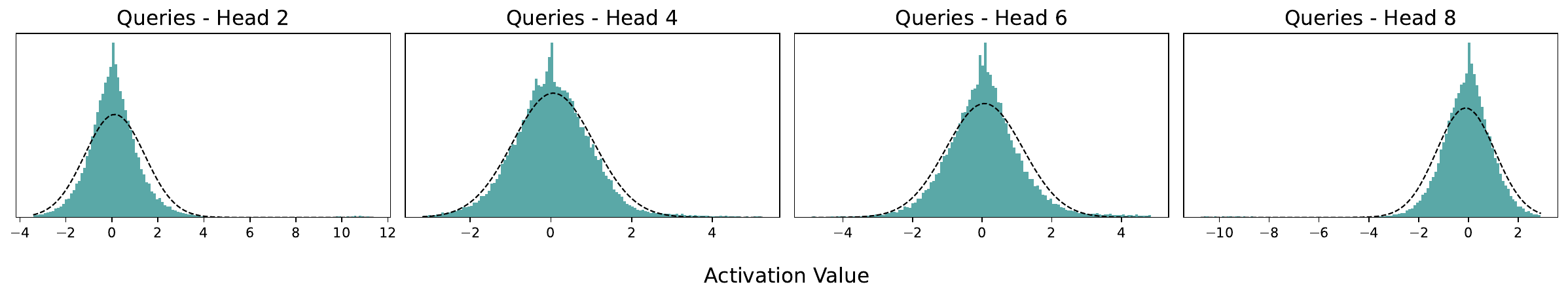}
        \caption{\qwen queries distributions.}
        \label{fig:qwen_q}
    \end{subfigure}
    \caption{Distributions of \qwen Hidden States and queries.}
    \label{fig:qwen_combined}
\end{figure}

\begin{figure}[t]
    \centering
    \begin{subfigure}[t]{\linewidth}
        \centering
        \includegraphics[width=\linewidth]{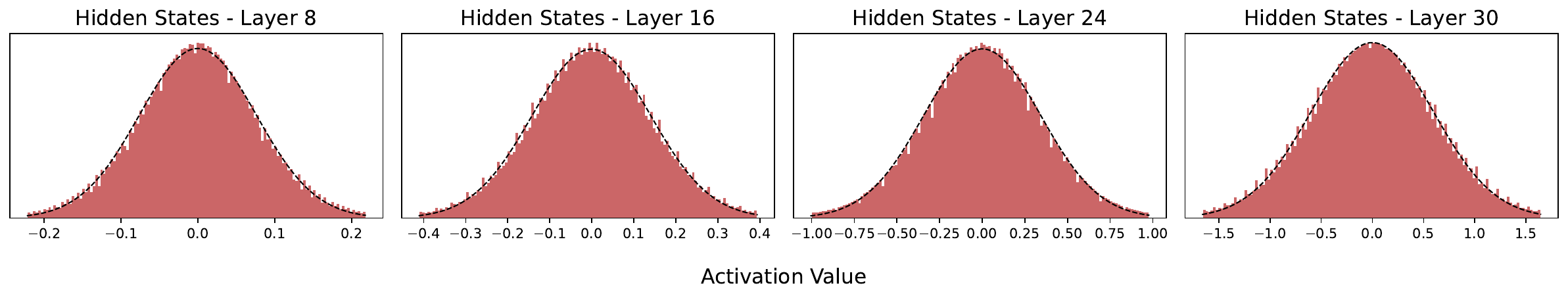}
        \caption{\llama hidden states distributions.}
        \label{fig:llama_hs}
    \end{subfigure}
    \begin{subfigure}[t]{\linewidth}
        \centering
        \includegraphics[width=\linewidth]{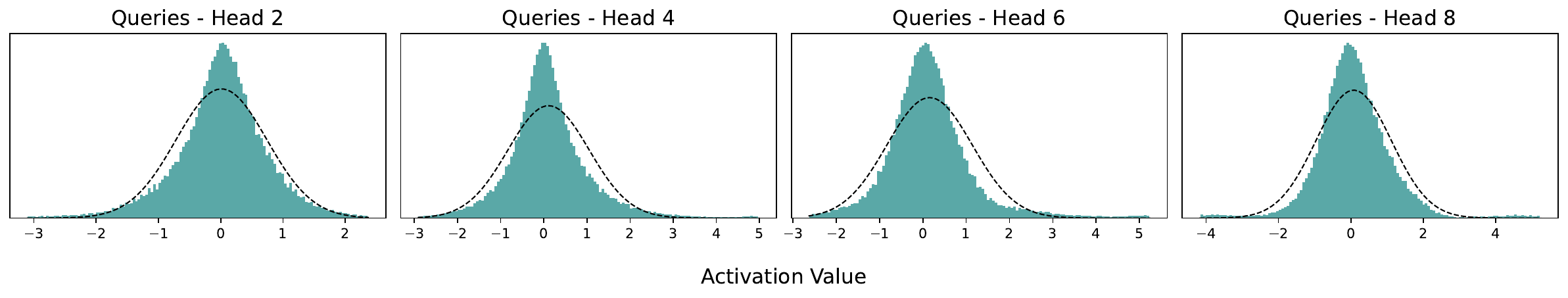}
        \caption{\llama queries distributions.}
        \label{fig:llama_q}
    \end{subfigure}
    \caption{Distributions of \llama hidden states and queries.}
    \label{fig:llama_combined}
\end{figure}

\begin{figure}[t]
    \centering
    \begin{subfigure}[t]{\linewidth}
        \centering
        \includegraphics[width=\linewidth]{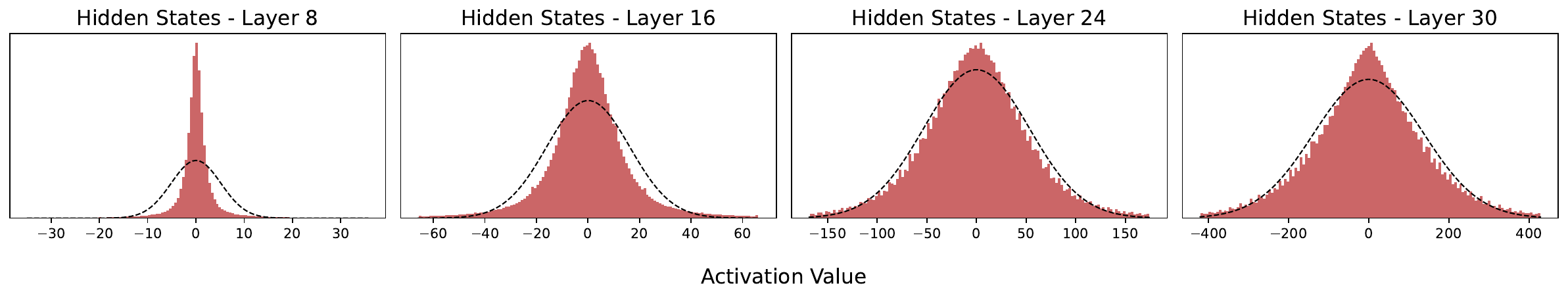}
        \caption{\gemma hidden states distributions}
        \label{fig:gemma_hs}
    \end{subfigure}
    \begin{subfigure}[t]{\linewidth}
        \centering
        \includegraphics[width=\linewidth]{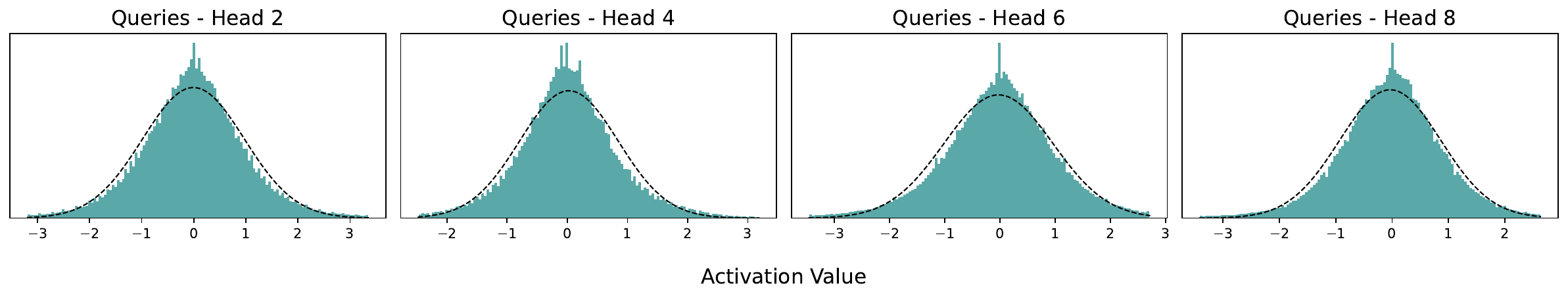}
        \caption{\gemma queries distributions.}
        \label{fig:gemma_q}
    \end{subfigure}
    \caption{Distributions of \gemma hidden states and queries.}
    \label{fig:gemma_combined}
\end{figure}

For \ours, we are interested in the hidden states before the MLP layers and the corresponding queries. Our study confirms that such activations are predominantly unimodal and can be approximated as Gaussian distributions, albeit with the presence of a few heavy-tailed outliers, as already found in ~\citet{streamingllm, massive}. In \Cref{fig:gemma_hs}, \Cref{fig:llama_hs}, and \Cref{fig:qwen_hs} we show hidden states and queries for different models. For our method, the distributional properties of queries are of particular importance, and we observe that queries maintain a clear Gaussian-like behaviour. This also applies to models with QK normalization, where the query projection is not guaranteed to be linear. The concentration of these activations around a central value and their Gaussian like shape provides the theoretical basis for \ours.

We stress that in this work, our goal is not to explain or investigate this property, but rather to leverage it for \kvcache compression.
\section{Expected Attention Score }
\label{app:abl_score}
\begin{figure}[t]
    \centering
    \begin{subfigure}[t]{0.31\linewidth}
        \centering
        \includegraphics[width=\linewidth]{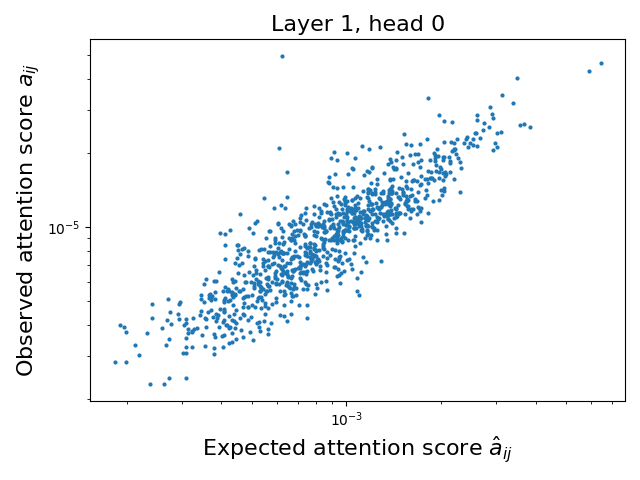}
    \end{subfigure}
    \begin{subfigure}[t]{0.31\linewidth}
        \centering
        \includegraphics[width=\linewidth]{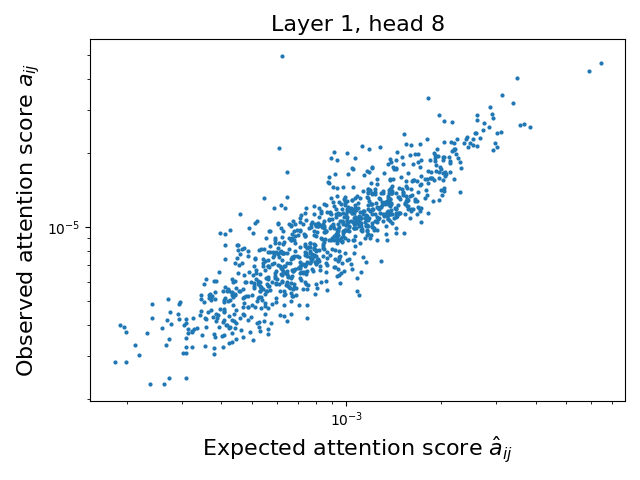}
    \end{subfigure}
    \begin{subfigure}[t]{0.31\linewidth}
        \centering
        \includegraphics[width=\linewidth]{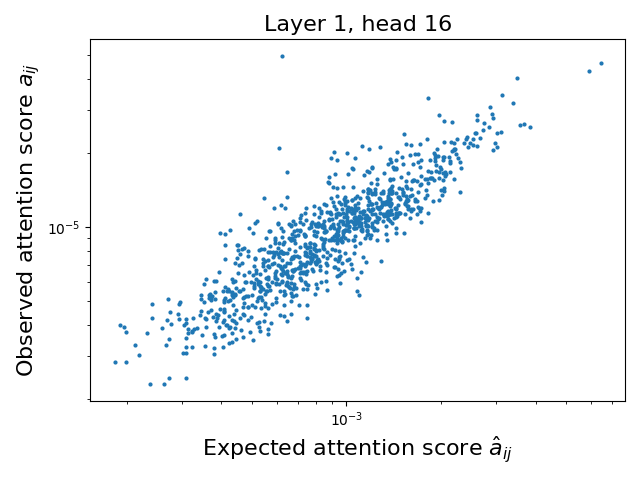}
    \end{subfigure}

    \begin{subfigure}[t]{0.31\linewidth}
        \centering
        \includegraphics[width=\linewidth]{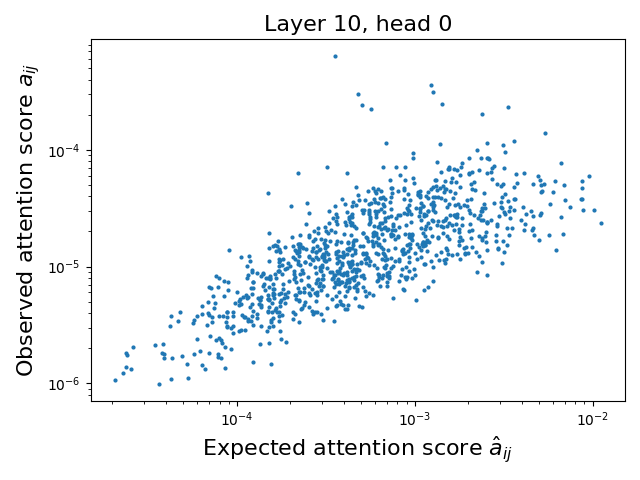}
    \end{subfigure}
    \begin{subfigure}[t]{0.31\linewidth}
        \centering
        \includegraphics[width=\linewidth]{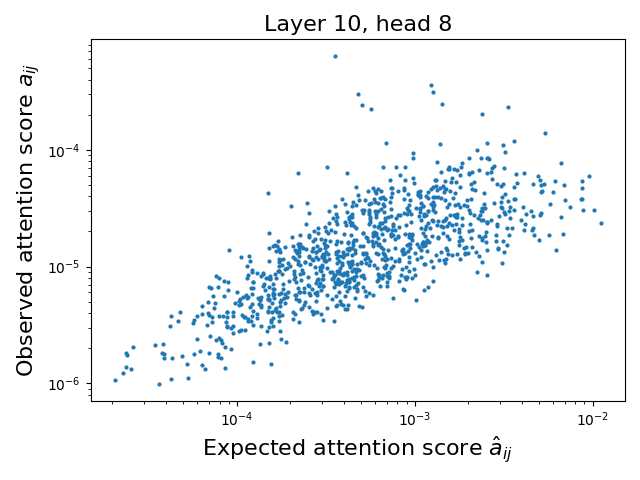}
    \end{subfigure}
    \begin{subfigure}[t]{0.31\linewidth}
        \centering
        \includegraphics[width=\linewidth]{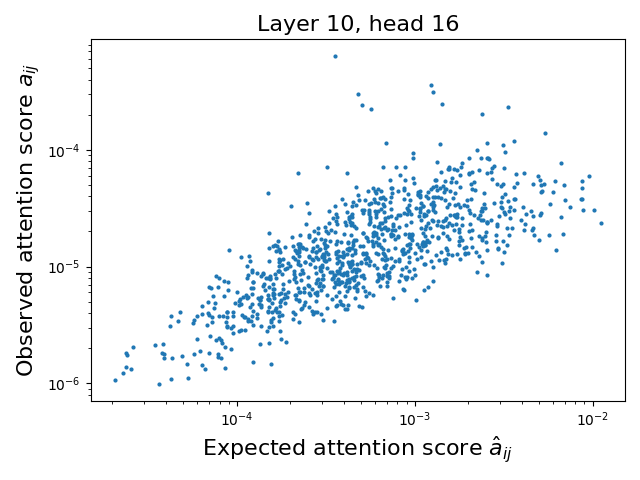}
    \end{subfigure}

    \begin{subfigure}[t]{0.31\linewidth}
        \centering
        \includegraphics[width=\linewidth]{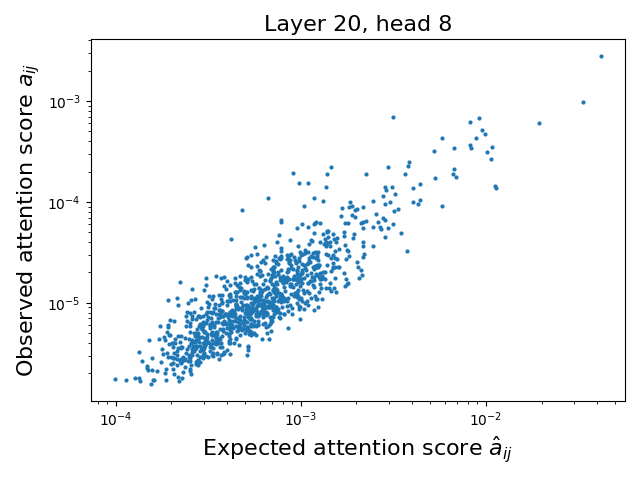}
    \end{subfigure}
    \begin{subfigure}[t]{0.31\linewidth}
        \centering
        \includegraphics[width=\linewidth]{figures/corr/l_20_h_8.png}
    \end{subfigure}
    \begin{subfigure}[t]{0.31\linewidth}
        \centering
        \includegraphics[width=\linewidth]{figures/corr/l_20_h_8.png}
    \end{subfigure}
    
    \caption{Correlation between attention score and expected attention score for \llama. We compute the expected attentions score on a sequence of 5K tokens, using the first 1K for statistics. A strong correlation exists between our attention score approximation and the observed attention score.}
    \label{fig:corrs}
\end{figure}

To empirically validate that the expected attention score is strongly correlated to the real model attention score, we plot the correlation between the observed attention and the expected attention score across different layers and heads. We use sequence of 5K tokens and use the first 1K tokens to compute the query statistics. We display the results in \Cref{fig:corrs}. We see that for different layers and attention heads, the expected attention score from Equation~\ref{eq:score} is strongly correlated to the original attention score.

\section{Additional Results}
\label{app:more_results}
In \Cref{tab:longbench} we show additional results on the LongBench dataset, averaged across all subsets.  

\begin{table}[t]
\centering
\caption{\ours outperforms most baselines on Longbench~\citep{longbench}. We show average score with increasing compression ratios across baselines.}
\label{tab:longbench}
\def\tablescale{0.95}
\scalebox{\tablescale}{
\begin{tabularx}{\textwidth}{ll*{5}{>{\centering\arraybackslash}X@{\hspace{8mm}}}>{\centering\arraybackslash}X}
    \toprule
    \multirow{2}{*}{\textbf{Model}} & \multirow{2}{*}{\textbf{Method}} & \multicolumn{6}{c}{\textbf{Longbench}} \\
    \cmidrule(lr){3-8}
    & & \textbf{0\%} & \textbf{10\%} & \textbf{25\%} & \textbf{50\%} & \textbf{75\%} & \textbf{90\%} \\
    \toprule      
    \multirow{4}{*}{\textit{\qwen}}  
    & Expected Attention   & $     \textbf{48.63}     $ & $   48.30      $ & $   \m{50.25}   $ & $    \m{50.1}    $ & $   \m{48.06}   $ & $   \m{39.71}   $ \\ 
    & TOVA                 & $     \textbf{48.63}     $ & $   48.41      $ & $   48.14   $ & $   46.49   $ & $   43.19   $ & $   37.21   $ \\ 
    & SnapKV               & $     \textbf{48.63}     $ & $   \m{48.40}    $ & $   47.85   $ & $   46.25   $ & $   42.42   $ & $   34.57   $ \\ 
    & KeyDiff              & $     \textbf{48.63}     $ & $   48.13      $ & $   46.23   $ & $   40.08   $ & $   29.42   $ & $   20.69   $ \\ 
    \midrule
    \multirow{4}{*}{\textit{\gemma}}
    & Expected Attention   & $     \textbf{51.04}     $ & $   \m{54.02}   $ & $   50.98   $ & $   47.51   $ & $   40.41   $ & $   32.67   $ \\ 
    & TOVA                 & $     \textbf{51.04}     $ & $   53.05   $ & $   \m{51.52}   $ & $    \m{50.7}    $ & $   \m{46.88}   $ & $   \m{40.45}   $ \\ 
    & SnapKV               & $     \textbf{51.04}     $ & $   51.83   $ & $   51.31   $ & $   48.14   $ & $   44.31   $ & $   34.97   $ \\ 
    & KeyDiff              & $     \textbf{51.04}     $ & $   51.64   $ & $   48.74   $ & $   42.15   $ & $   33.68   $ & $   23.46   $ \\ 
    \midrule
    \multirow{4}{*}{\textit{\llama}}
    & Expected Attention   & $     \textbf{46.42}     $ & $   \m{46.59}   $ & $   \m{46.8}    $ & $   \m{47.91}   $ & $   \m{44.04}   $ & $   33.97   $ \\ 
    & TOVA                 & $     \textbf{46.42}     $ & $   46.22      $ & $   45.62   $ & $   44.13   $ & $    40.5    $ & $   34.77   $ \\ 
    & SnapKV               & $     \textbf{46.42}     $ & $   46.56      $ & $   46.07   $ & $   45.07   $ & $   41.24   $ & $   32.55   $ \\ 
    & KeyDiff              & $     \textbf{46.42}     $ & $   46.45      $ & $   48.01   $ & $    46.9    $ & $   42.24   $ & $   \m{35.51}   $ \\ 
    \bottomrule
\end{tabularx}
}
\end{table}

\begin{figure}
    \centering
    \includegraphics[width=0.7\linewidth]{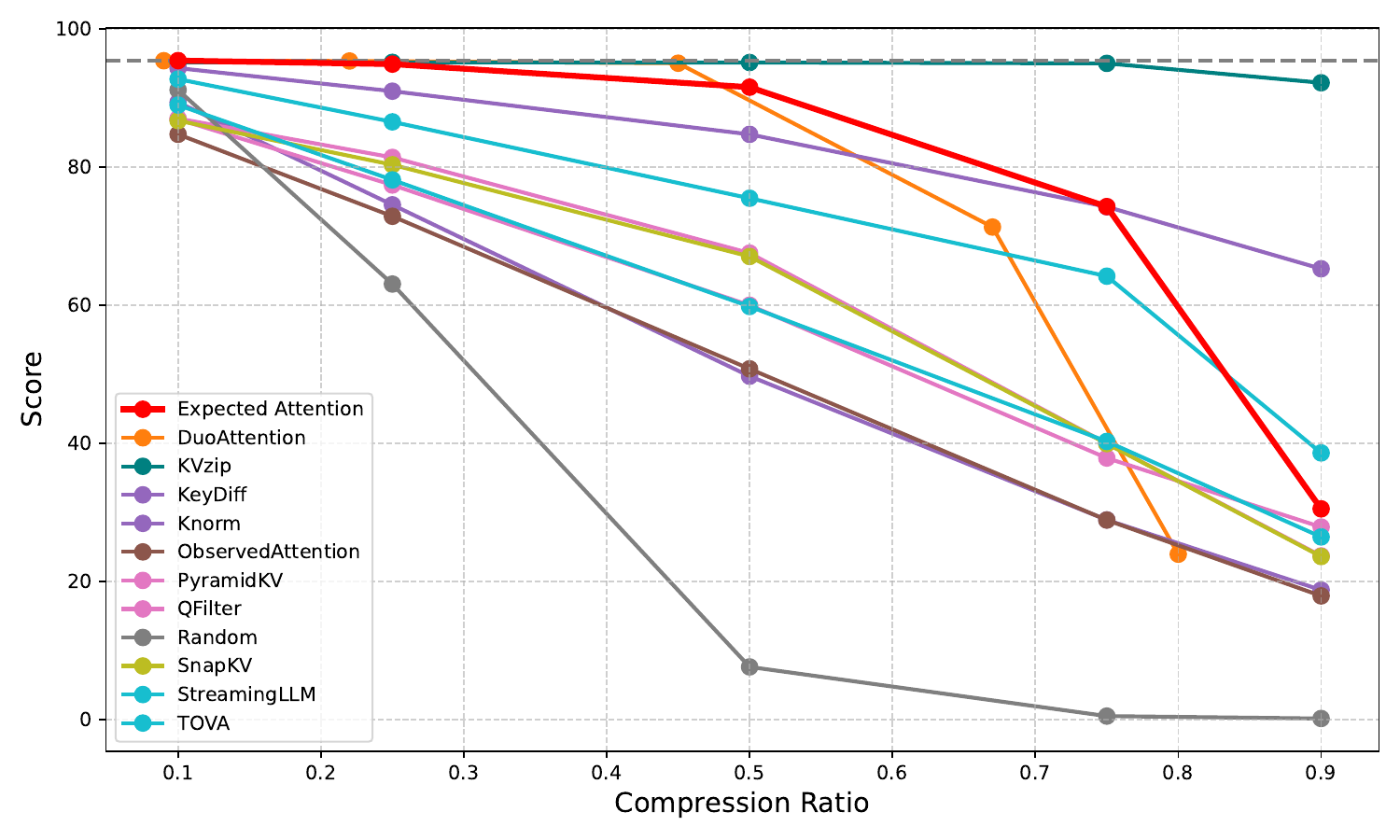}
    \caption{Initial experiments on Ruler 4K to select the best baselines. We did not use KVZip as it requires two forward passes and increases latency significantly.}
    \label{fig:leaderboard}
\end{figure}

\paragraph{Ruler}
In order to select the most competitive baselines we performed an initial search on 15+ methods on Ruler. We selected the best performing ones as displayed in \Cref{fig:leaderboard}. We did not include KVZip~\citep{kim2025kvzipqueryagnostickvcache} despite achieving a high score as it needs two forward passes, therefore implying a higher cost FLOPs that is double as much as the other baselines. 

%
%
%
%

\end{document}